\documentclass[twoside]{article}

\usepackage{xcolor}
\usepackage{graphicx}
\usepackage{amsfonts}
\usepackage{float}
\usepackage{siunitx}
\usepackage{multirow}
\usepackage{booktabs}
\usepackage{url}
\usepackage{mathtools}

\DeclareMathOperator*{\argmin}{arg\,min}
\usepackage[accepted]{aistats2025}
\usepackage[round]{natbib}

\usepackage[hidelinks]{hyperref}

\begin{document}

% If your paper is accepted and the title of your paper is very long,
% the style will print as headings an error message. Use the following
% command to supply a shorter title of your paper so that it can be
% used as headings.
%
%\runningtitle{I use this title instead because the last one was very long}

% If your paper is accepted and the number of authors is large, the
% style will print as headings an error message. Use the following
% command to supply a shorter version of the authors names so that
% they can be used as headings (for example, use only the surnames)
%
%\runningauthor{Surname 1, Surname 2, Surname 3, ...., Surname n}

\twocolumn[

\aistatstitle{On Local Posterior Structure in Deep Ensembles}

\aistatsauthor{ Mikkel Jordahn$^*$ \And Jonas Vestergaard Jensen$^*$ \And  Mikkel N. Schmidt \And Michael Riis Andersen }

\aistatsaddress{  Technical University of Denmark \\  $^*$Shared first authorship \\ \texttt{\{mikkjo,jovje\}@dtu.dk}} ]

\begin{abstract}
Bayesian Neural Networks (BNNs) often improve model calibration and predictive uncertainty quantification compared to point estimators such as maximum-a-posteriori (MAP). Similarly, deep ensembles (DEs) are also known to improve calibration, and therefore, it is natural to hypothesize that deep ensembles of BNNs (DE-BNNs) should provide even further improvements. In this work, we systematically investigate this across a number of datasets, neural network architectures, and BNN approximation methods and surprisingly find that when the ensembles grow large enough, DEs consistently outperform DE-BNNs on in-distribution data. To shine light on this observation, we conduct several sensitivity and ablation studies. Moreover, we show that even though DE-BNNs outperform DEs on out-of-distribution metrics, this comes at the cost of decreased in-distribution performance. As a final contribution, we open-source the large pool of trained models to facilitate further research on this topic.
\end{abstract}

\section{INTRODUCTION}

Regardless of neural networks' immense popularity and successes, they are still seeing limited application in areas where uncertainty quantification and model calibration is of high importance such as in healthcare \citep{kompa2021communication,seoni2023uqhealthcare}. This is in large part due to the widely known phenomenon that modern neural networks are often overconfident in their predictions \citep{nguyen2015highconfnn,guo2017calibmodernnn}. Bayesian inference applied to neural networks, i.e., Bayesian deep learning, is one area of research that in theory brings the promise of remedying poor model calibration and uncertainty quantification \citep{papamarkou24BayesianDeepLearning}.

In recent years, many different approximate Bayesian inference methods have been proposed for neural networks \citep{mackay1992bayesian,2015_BayesByBack, gal2016dropout,maddox2019simple,dusenberry2020efficient,daxberger2021laplace, 2024_VBLL}. Many of these methods have been shown to improve both in- and out-of-distribution uncertainty quantification. However, the majority of these only describe the structure around a single mode of the posterior even though neural networks posteriors are known to be highly multimodal. On the contrary, deep ensembles (DEs) \citep{1990_LKAIDE, lakshminarayanan2017simple} capture several modes of the posterior and can be interpreted as approximate Bayesian inference \citep{wilson2020bayesian,DAngelo2021RepulsiveDE}, yet do not capture the local structure around these modes. Nevertheless, DEs are often found to perform better than approximate inference methods that describe the local structure of a single mode. Based on the improvements provided by unimodal approximations (e.g. the Laplace approximations) and DEs separately, it is a natural hypothesis that equipping the modes of DEs with local posterior structure should further improve their performance.

In this work, we systematically investigate the effect of including local posterior structure in DEs (DE-BNNs) through approximate inference methods, namely stochastic weight averaging Gaussian (SWAG) \citep{maddox2019simple}, last-layer Laplace approximation (LLLA) \citep{daxberger2021laplace}, and LLLA refined by normalizing flows \citep{kristiadi2022posterior}. In contrast to previous works, we investigate this across different datasets, neural network architectures and ensemble sizes, and surprisingly find that as we increase the number of members in the ensembles, the DEs consistently outperform the DE-BNNs on in-distribution metrics. Interestingly, we also find that even though the DE-BNNs perform slightly better on out-of-distribution metrics, this often comes at the cost of worse in-distribution performance. We explore this behaviour through a range of sensitivity and ablation studies. Our contributions can be summarized as the following:

\begin{enumerate}
    \item We show that DEs generally outperform DE-BNNs across a number of datasets, neural networks architectures and approximate BNN methods for large ensemble sizes.
    \item We conduct a number of sensitivity and ablation studies to explain the different predictive performance between BNN and DE-BNNs.
    \item We show that increased out-of-distribution performance in DE-BNNs often comes at an in-distribution performance cost.
    \item We open-source a large class of trained BNNs for further analysis\footnote{Code and models via \url{https://github.com/jonasvj/OnLocalPosteriorStructureInDeepEnsembles}}.
\end{enumerate}

Based on these contributions and experiments, we present advice for practitioners on model choice.

\section{RELATED WORK}

There exists a wide range of works in Bayesian deep learning concerned with approximating the intractable posterior over neural networks. These include stochastic variational inference (SVI) methods \citep{2015_BayesByBack, dusenberry2020efficient, variational2024shen,2024_VBLL}, Laplace approximations \citep{mackay1992bayesian,Daxberger2020BayesianDL,daxberger2021laplace}, Monte Carlo dropout \citep{gal2016dropout}, and stochastic gradient based approximation methods \citep{stephan2017stochastic,maddox2019simple}, all of which have been shown to improve uncertainty quantification and predictive performance over \textit{maximum a-posterior} (MAP) estimators. Common for many of these methods is that they only approximate the posterior distribution around a single mode, which is a crude approximation considering the wide-spread knowledge that loss-landscapes of neural networks are highly multimodal.

Hamiltonian Monte Carlo (HMC) sampling is the golden standard for approximating the posterior distribution of neural networks \citep{neal1996HMCBNNs}, and can in principle explore multiple modes, but it is highly impractical to run due to computational cost. The largest HMC based BNN experiment ever done was completed in \citet{izmailov2021bayesian} where some chains took more than \textit{60 million epochs} to converge on CIFAR-10---and even then, some have later questioned whether the HMC chains are fully exploring the true posterior \citep{sharma2023dobayesian}. Deep ensembles (DEs) \citep{1990_LKAIDE, lakshminarayanan2017simple}, on the other hand, are straightforward to implement and train, and afford large performance increases even compared to the aforementioned BNN methods. However, DEs are still only point estimates at each mode, and it seems intuitive that including local posterior structure in the DEs should further improve them. 

\citet{wilson2020bayesian} and \citet{Eschenhagen2021MixturesOL} have previously experimented with including local posterior structure in DEs using SWAG \citep{maddox2019simple} and LLLA \citep{daxberger2021laplace}. In our work, we similarly investigate SWAG and LLLA and additionally the LLLA refined by normalizing flows \citep{kristiadi2022posterior} for including local posterior structure in DEs but on a larger class of problems and architectures. While our results do not conflict with \citet{wilson2020bayesian} and \citet{Eschenhagen2021MixturesOL}, we present additional findings that shine a new and different light on the current state of DE-BNNs. Firstly, we show that DEs generally perform on-par with or better than DE-BNNs on in-distribution metrics when the ensembles grow large enough. Secondly, we show that increased out-of-distribution performance observed in DE-BNNs comes at a cost of in-distribution performance.

\section{BACKGROUND}
\label{sec:background}

In Bayesian inference, the goal is to compute the posterior distribution over model parameters $\theta \in \mathbb{R}^D$ given data $\mathcal{D} = \{(x_n, y_n)\}_{n=1}^N$
\begin{equation}
\label{eq:bayes}
    p({\theta|\mathcal{D})} = \frac{p(\mathcal{D}|\theta)p(\theta)}{p(\mathcal{D})} \,,
\end{equation}
where $p(\mathcal{D}|\theta)$ is the likelihood, $p(\theta)$ is a prior distribution over the parameters, and $p(\mathcal{D})$ is the \textit{marginal likelihood}. Given the posterior distribution $p(\theta|\mathcal{D})$, we can make predictions for a new data point $x^*$ via the posterior predictive distribution
\begin{equation}
\label{eq:predictive}
    p(y^*|x^*,\mathcal{D})=\int_{\mathbb{R}^D} p(y^*, |x^*,\theta)p(\theta|\mathcal{D})\,d\theta \,,
\end{equation}

through marginalization of the model parameters $\theta$.

In Bayesian deep learning both Eq.~\eqref{eq:bayes} and \eqref{eq:predictive} are intractable, and we therefore resort to an approximation $q(\theta) \approx p({\theta|\mathcal{D})}$ and further estimate the predictive distribution with Monte Carlo (MC) sampling
\begin{equation}
\label{eq:predictive_mc}
    p(y^*|x^*,\mathcal{D}) \approx \frac{1}{S}\sum_{s=1}^S p(y^*|x^*,\theta^{(s)})\,, \quad\theta^{(s)} \sim q(\theta)\,.
\end{equation}

Deep Ensembles (DEs) are a collection of $K$ independently trained neural networks. If the loss function is chosen as $-\log p(\mathcal{D},\theta)$ and the likelihood has the form $p(\mathcal{D}|f_{\theta^{(k)}})$ where $f_{\theta^{(k)}}$ is the $k$th neural network, then the DE can be seen as a collection of $K$ MAP estimates $\{\theta_{\mathrm{MAP}}^{(k)}\}_{k=1}^K$, i.e., $K$ (local) maximizers of Eq.~\eqref{eq:bayes} w.r.t.\ $\theta$. In the DE, predictions are made according to
\begin{equation}
\label{eq:deep_ensemble}
    p(y^*|x^*, \mathcal{D}) \approx \frac{1}{K}\sum_k p(y^*|x^*, \theta_{\mathrm{MAP}}^{(k)}) \,.
\end{equation}
Noting the similarity between Eq.~\eqref{eq:predictive_mc} and \eqref{eq:deep_ensemble}, we can view DEs as a direct approximation of Eq.~\eqref{eq:predictive}, or equivalently as an approximate posterior with the form
\begin{equation}
\label{eq:de_posteiror}
    q(\theta) = \sum_{k=1}^K \pi_k \delta(\theta-\theta_{\mathrm{MAP}}^{(k)})\,,
\end{equation}
where $\pi_k=\frac{1}{K}\ \forall k$ and $\delta$ is Dirac's delta distribution. 

In this work, we equip DEs with local posterior structure around the MAP estimates by considering approximate Bayesian inference methods that can be computed \textit{post-hoc} from a MAP estimate. Our approximate posteriors therefore take the form
\begin{equation}
\label{eq:approximate_posterior}
    q(\theta)=\sum_{k=1}^K \pi_k q_k(\theta | \theta_{\mathrm{MAP}}^{(k)}) \,.
\end{equation}
where $\pi_k \in [0,1]\ \forall k$ and $\sum_{k=1}^K\pi_k=1$.
Next, we discuss our choices for the local distributions $q_k$.

\subsection{Last-Layer Laplace Approximation}
The Laplace Approximation (LA) is a local Gaussian approximation to $p(\theta|\mathcal{D})$ centered at a MAP estimate $\theta_{\mathrm{MAP}}$. The LA is given by 
\begin{equation}
    q(\theta|\theta_{\mathrm{MAP}}) = \mathcal{N}(\theta|\theta_{\mathrm{MAP}}, \mathcal{H}^{-1}),
\end{equation}
 where $\mathcal{H}$ is the Hessian of the negative joint distribution, i.e.,
\begin{equation}
     \mathcal{H}=-\nabla^2\log p(\mathcal{D},\theta)\Bigr|_{\theta=\theta_{\mathrm{MAP}}}.
\end{equation}

The size of the Hessian $\mathcal{H}$ scales quadratically in the number of network parameters $D$ and it is therefore often necessary to approximate $\mathcal{H}$, e.g. by imposing constraints on $\mathcal{H}$, in a deep learning context \citep{daxberger2021laplace}. In this work, we consider the last-layer Laplace approximation (LLLA) \citep{kristiadi2020being} in which the LA is only constructed for the parameters in the last layer of the neural network and all other parameters are fixed to their MAP estimate, i.e.,
\begin{equation}
\label{eq:llla}
    q(\theta | \theta_{\mathrm{MAP}}) = \delta(\theta^{-L}\!-\!\theta_{\mathrm{MAP}}^{-L})\mathcal{N}\big(\theta^L|\theta_{\mathrm{MAP}}^L, (\mathcal{H}^L)^{-1}\big),
\end{equation}
where the superscript $L$ denotes parameters in the last layer and $-L$ indexes the parameters in all earlier layers such that $\theta = \theta^{-L} \cup \theta^L$.

\subsection{Normalizing Flows for the Last-Layer Laplace Approximation}
Since neural network posteriors are complex and non-Gaussian \citep{garipov2018loss,izmailov2021bayesian}, we leverage normalizing flows \citep{rezende2015variational} to refine the Gaussian LLLA, as proposed by \citet{kristiadi2022posterior}.

A normalizing flow creates a random variable $\theta \in \mathbb{R}^D$ with a complex probability density function by transforming a random variable $\theta_0$ with distribution $q(\theta_0)$ using a diffeomorphism $F_{\phi}:\mathbb{R}^D \to \mathbb{R}^D$.
$F_{\phi}$ is constructed by composing $T$ simple diffeomorphic transformations, i.e., $F_{\phi}= F_{\phi_T} \circ F_{\phi_{T-1}} \circ \cdots \circ F_{\phi_{1}}$ with $\phi=\{\phi_t\}_{t=1}^T$, which ensures that the composite transformation $F_{\phi}$ is also a diffeomorphism \citep{papamakarios2021normalizing}.
Samples of $\theta$ are obtained by first sampling $\theta_{0}$ and then applying $F_{\phi}$, i.e.,
\begin{equation}
    \theta = F_{\phi}(\theta_0) = F_{\phi_T}\left(\cdots F_{\phi_1}\left(\theta_0\right)\right)\,,\quad \theta_0 \sim q(\theta_0) \,,
\end{equation}
and densities are computed with the change-of-variables formula

\begin{equation}
\begin{split}
\label{eq:flow_dist}
    q_{\phi}(\theta)
    &= q(\theta_0)\vert \mathrm{det} J_{F_{\phi}}(\theta_0)|^{-1}
    \\
    &= q(\theta_0)\left \vert \prod_{t=1}^T \mathrm{det} J_{F_{\phi_t}}(\theta_{t-1})\right|^{-1} \,,
\end{split}
\end{equation}
with $\theta_0 = F_{\phi}^{-1}(\theta)$ and where $J_{F_{\phi_t}}$ is the Jacobian of $F_{\phi_t}$ and $\theta_t = F_{\phi_t}(\theta_{t-1})$ for $t=1,\dots,T$ with $\theta=\theta_T$.

When refining the LLLA with normalizing flows, we use the LLLA given in Eq.~\eqref{eq:llla} as the base distribution $q(\theta_0)$ and minimize the (reverse) Kullback-Leibler (KL) divergence between $q_{\phi}(\theta)$ given in Eq.~\eqref{eq:flow_dist} and the true posterior $p(\theta|\mathcal{D})$ by optimizing the variational parameters $\phi$, i.e.,
\begin{equation}
    \phi^* = \argmin_{\phi} \mathbb{KL}[q_{\phi}\Vert p]
\end{equation}
In practice, this is done by maximizing the evidence lower bound (ELBO) w.r.t.\ $\phi$. 

\subsection{Constant Stochastic Gradient Descent}

The analysis of \citet{stephan2017stochastic} shows that the iterates from stochastic gradient descent (SGD) with a constant learning rate can be seen as samples from a Gaussian approximate posterior. SWA-Gaussian (SWAG) is a practical algorithm for estimating the mean and covariance of the SGD iterates' stationary distribution. Denoting the SGD iterate after epoch $m$ as $\theta_m$, SWAG estimates the mean after $M$ epochs on the objective $-\log p(\theta,\mathcal{D})$ as
\begin{equation}
    \theta_{\mathrm{SWA}} = \frac{1}{M}\sum_{m=1}^M \theta_m \,,
\end{equation}
which is the stochastic weight averaging (SWA) \citep{izmailov2018} estimate of $\theta$. The covariance matrix in SWAG is composed of a diagonal component
\begin{equation}
    \Sigma_{\mathrm{diag}} = \mathrm{diag}(\bar{\theta^2} - \theta_{\mathrm{SWA}}^2)\,,\quad \bar{\theta^2} = \frac{1}{M}\sum_{m=1}^M \theta_m^2\,,
\end{equation}
(where the squares are element-wise) and a low-rank approximation of rank $R$ using the $R$ last iterates
\begin{equation}
\label{eq:low_rank}
    \Sigma_{\textrm{low-rank}} = \frac{1}{R-1}\sum_{\mathclap{\substack{m\in\\\{M-R+1,\dots,M\}}}} (\theta_m - \bar{\theta}_R)(\theta_m - \bar{\theta}_R)^{\top}\,,
\end{equation}
where $\bar{\theta}_R$ is the mean of the iterates from the final $R$ epochs.
The SWAG approximate posterior therefore takes the form
\begin{equation}
    q(\theta) = \mathcal{N}\left(\theta\middle|\theta_{\mathrm{SWA}},\frac{1}{2}(\Sigma_{\mathrm{diag}} + \Sigma_{\textrm{low-rank}})\right)\,.
\end{equation}
In our experiments we initialize constant-learning-rate SGD from a MAP estimate $\theta_{\mathrm{MAP}}$ and the SWAG posterior therefore has an implicit dependence on $\theta_{\mathrm{MAP}}$.

\begin{figure*}[t!]
    \centering
    \vspace{-0.4cm}
    
{%
        \includegraphics[width=1\linewidth]{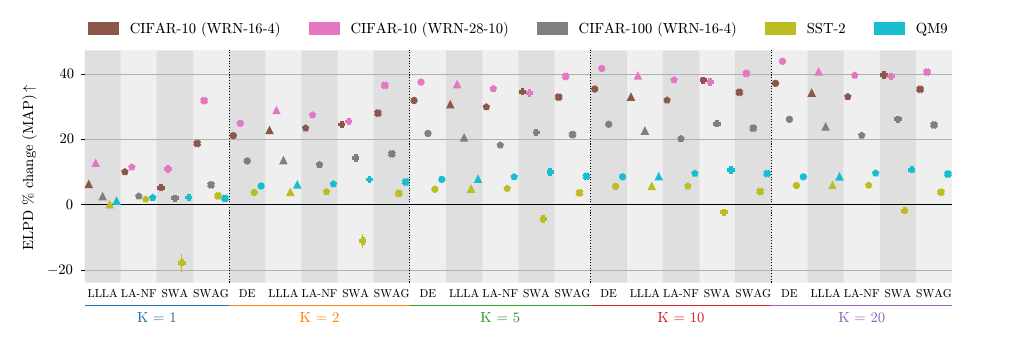}%
    }
    \vspace{-0.4cm}
    
{%
    \includegraphics[width=1\linewidth, trim={0 0 0 20}, clip]{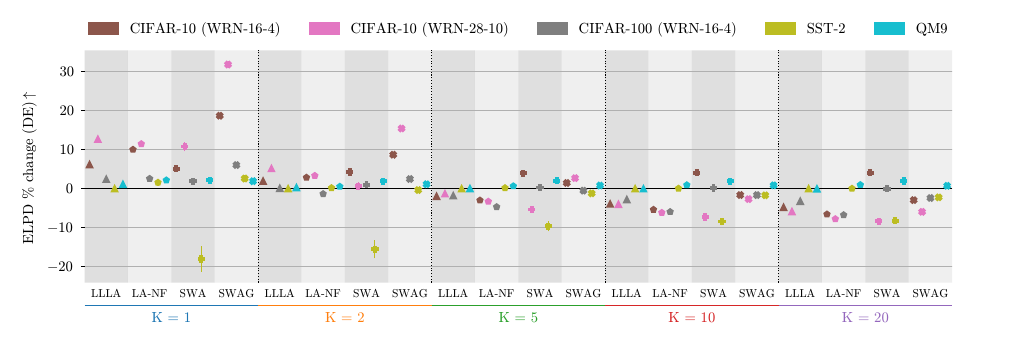}%
    }
    \caption{Top: Test ELPD change in percentage of DEs and DE-BNNs relative to MAP estimates with $K=1$. DEs are not included here for $K=1$ as these corresponds to the MAP estimates. Bottom: Test ELPD change in percentage of DE-BNNs relative to DEs. All points located below $y=0$ indicate that the DE of the same ensemble size outperform the equivalently sized DE-BNN method.}

    \label{fig:ELPD_metrics}
\end{figure*}

\section{EXPERIMENTAL SETUP}
\label{sec:experimental_setup}
We design and conduct a series of experiments to investigate the hypothesis that enriching deep ensembles with local posterior structure improves the predictive performance. We are particularly interested in uncertainty quantification and calibration, and therefore, we focus on the expected log predictive density (ELPD) \citep{gelman2014understanding} as evaluation metric. However, we also provide results for both accuracy and expected calibration error (ECE) \citep{guo2017calibmodernnn}. In regression settings we report the negative mean absolute error (N-MAE) in place of accuracy. All results are presented as means $\pm$ 2 standard errors across seeds.

\subsection{Datasets and Models}
We conduct experiments on four real and diverse datasets: CIFAR-10 and CIFAR-100 image classification datasets \citep{krizhevsky2009learning}, the sentiment classification dataset SST-2 \citep{socher2013recursive}, and the molecular property prediction dataset QM9 \citep{wu2018moleculenet}, using the zero-point energy $U_0$ as target. We also conduct out-of-distribution (OOD) experiments. For the image classification models we use SVHN \citep{netzer2011reading} and CIFAR-100 data as OOD data for CIFAR-10 models and SVHN and CIFAR-10 data as OOD data for CIFAR-100 models. For SST-2 models we use the Yahoo Finance News Sentences dataset \citep{yahoo} and for QM9 models we use the Tencent Alchemy dataset \citep{chen2019alchemy}.

For the classification tasks, we use a categorical likelihood of the form $p(\mathcal{D} | \theta) = \prod_{n=1}^N \mathrm{cat}\big(y_n\big|f_{\theta}(x_n)\big)$
where $f_{\theta}(x)$ is a neural network that predicts the class probabilities for a given $x$. We use the WRN-16-4 and WRN-28-10 architectures \citep{zagoruyko2016wide} for CIFAR-10 and WRN-16-4 for CIFAR-100. For the SST-2 dataset, we use a small BERT architecture \citep{devlin2019bert} (see Appendix \ref{app:training_details}). For the WRN models we use filter response normalization (FRN) \citep{singh2020filter} due to issues with batch normalization and BNNs (see Appendix \ref{app:FRN_BN}).

The molecular property prediction task from the QM9 dataset is a regression problem, and we model this using a hetereoscedastic Gaussian likelihood $ p(\mathcal{D} | \theta)= \prod_{n=1}^N \mathcal{N}(y_n | f_{\theta}^{\mu}(x_n), f_{\theta}^{\sigma^2}(x_n))$, where $f_{\theta}$ is neural network that both predicts the mean and variance of $y_n$. For $f_{\theta}$, we adapt the PaiNN graph neural network from \cite{schutt2021equivariant} to heteroscedastic regression.

\subsection{Model Training}

We train each model 30 times with different initializations, yielding 30 MAP estimates of the model parameters $\theta$ for each pair of dataset and model. We then compute each of the three post-hoc approximate posteriors (SWAG, LLLA, LA-NF) for each MAP estimate.  Finally, we construct DEs and DE-BNNs with Eq.~\eqref{eq:de_posteiror} and \eqref{eq:approximate_posterior} by randomly generating 30 model combinations without replacement for each ensemble size $K \in \{2, 5, 10, 20\}$.

The MAP estimates are obtained by minimizing the negative log-likelihood using weight decay as an implicit Gaussian-like prior. The full details of training the MAP estimates are given in Appendix \ref{app:training_details}. For the BERT models, we do not leverage pre-training as the computational cost would be prohibitively expensive.

For the post-hoc approximate posteriors, we carefully tune the variance of an explicit zero-mean isotropic Gaussian prior for both the LLLA and LA-NF and the constant learning rate for SWAG using the validation ELPD as selection criterion. We use $S=100$ samples to compute the predictive distribution in Eq.~\eqref{eq:predictive_mc} during hyperparameter selection and $S=200$ samples during model evaluation on the test set. We use $M=R=100$ in the SWAG approximation for all models except WRN-28-10 where we use $M=R=25$. We only use a diagonal Hessian in the LLLA for WRN-28-10 and the BERT models, but otherwise use a full Hessian. For the LA-NF we use $T\in\{1,5,10,30\}$ radial flows \citep{rezende2015variational}. Only results for $T=10$ are included in the main text, and the remaining results can be found in Appendix~\ref{app:additional_tables}. Full details of computing the approximate posteriors are given in Appendix \ref{app:training_details}.

Finally, we stratify the posterior samples across the $K$ ensemble components and use uniform weights $\pi_k=\frac{1}{K}$ in Eq.~\eqref{eq:approximate_posterior}. We investigate the effect of these choices in Section \ref{sec:sensitivity}.

We also experiment with the Improved Variational Online Newton (IVON) approximate inference method proposed in \citet{variational2024shen} and Monte Carlo dropout (MCDO) as proposed in \citet{gal2016dropout}, but refer to Appendix \ref{app:training_details} for training details. We only report results for these methods in Appendices \ref{app:additional_tables} and \ref{app:additional_figures} as these methods are not post hoc BNN methods, meaning that we cannot decouple the effect of including local posterior structure from the mode-finding (training) procedure as we can in the other post-hoc methods. However, the picture for these methods are the same as with the aforementioned methods, with the exception being MCDO on SST-2 and QM9. We discuss these results more in detail in Appendix \ref{app:IVON_MCDO_Discussion}.

\begin{table*}[h]
\centering
\caption{Average Rankings across Datasets. Ranks are computed for each seed individually yielding ranks 1-5.} 
\resizebox{\textwidth}{!}{
\begin{tabular}{lrrrrrrrrrrrrrrr}
\toprule
& \multicolumn{3}{c}{K = 1} & \multicolumn{3}{c}{K = 2} & \multicolumn{3}{c}{K = 5} & \multicolumn{3}{c}{K = 10} & \multicolumn{3}{c}{K = 20} \\
\cmidrule(lr){2-4} \cmidrule(lr){5-7} \cmidrule(lr){8-10} \cmidrule(lr){11-13} \cmidrule(lr){14-16}
Inference & Acc. & ELPD & ECE & Acc. & ELPD & ECE & Acc. & ELPD & ECE & Acc. & ELPD & ECE & Acc. & ELPD & ECE \\

\midrule
DE & 3.9 & 4.6 & 4.1 & 3.6 & 4.1 & 2.8 & 3.1 & 2.7 & \textbf{1.9} & 3.0 & \textbf{2.2} & \textbf{2.0} & 2.9 & \textbf{2.2} & \textbf{2.2} \\
SWA & 2.2 & 3.5 & 4.1 & 2.2 & 2.9 & 3.8 & \textbf{2.7} & 2.6 & 2.7 & \textbf{2.8} & 2.6 & 2.3 & \textbf{2.8} & 2.6 & 2.3 \\
SWAG & \textbf{1.6} & \textbf{1.4} & 2.2 & \textbf{2.0} & \textbf{1.7} & 2.9 & 2.8 & \textbf{2.4} & 3.4 & 3.1 & 2.9 & 3.4 & 3.2 & 3.1 & 3.2 \\
LLLA & 4.1 & 3.1 & 2.8 & 3.8 & 3.1 & \textbf{2.6} & 3.4 & 3.5 & 3.3 & 3.2 & 3.5 & 3.5 & 3.2 & 3.4 & 3.7 \\
LA-NF & 3.3 & 2.4 & \textbf{1.8} & 3.3 & 3.2 & 2.9 & 3.0 & 3.8 & 3.7 & 2.9 & 3.7 & 3.7 & 3.0 & 3.6 & 3.6 \\
\bottomrule
\end{tabular}

\label{tab:rankings}
}
\end{table*}

\section{EXPERIMENTS \& RESULTS}
\label{sec:results}

\subsection{In-Distribution Tests}

First, we investigate the in-distribution (ID) performance by evaluating ELPD on test data.

Fig.~\ref{fig:ELPD_metrics} (top) shows the percentage change in ELPD for all combinations of models, datasets and inference methods in comparison to MAP estimated models with $K=1$. Here we, unsurprisingly, observe that across all methods and datasets, increasing the ensemble size $K$ significantly improves performance. Secondly, it is seen that the BNN methods consistently yield significant performance increases for $K=1$, except for SWA on the SST-2 dataset. We hypothesize that SWA performs poorly on SST-2 because the BERT models are trained in a low data regime without pretraining, which makes the MAP estimators highly prone to overfitting on this problem which is then exacerbated with further (SWA) gradient steps.

In Fig.~\ref{fig:ELPD_metrics} (bottom) we show the percentage change in ELPD for all combinations of models, datasets and inference methods in comparison to DEs with the same number of members. Here we see that already when $K=2$, some of the DE-BNN methods perform only on-par or worse than the DE version with the same number of members. This trend continues as $K$ is increased to 20 as almost all DE-BNNs are outperformed by DEs of the same size. It is only on the QM9 dataset where some DE-BNNs marginally outperform their DE versions.

We also note that although SWA-based DEs outperform the MAP-based DEs in several cases, these models are also only point estimates. Furthermore, SWA-based DEs are in several cases better than SWAG-based DE-BNNs (e.g., on the QM9 dataset) which points to a detrimental effect of local posterior structure since SWAG is the SWA solution equipped with local posterior structure.

To further underline that DEs generally outperform DE-BNNs for large ensemble sizes, we refer to Table~\ref{tab:rankings} which shows that when we rank each inference method from 1 to 5 for each seed, dataset, and $K$, DEs rank the highest both for $K=10$ and $K=20$ on ELPD and ECE. SWA, the other point estimate-based method, comes in a close second, which further attests that local posterior structure is not necessarily beneficial. For all ensembles $K>1$, the test accuracy is similar across inference methods and ensemble sizes. For detailed metrics including a version of Fig.~\ref{fig:ELPD_metrics} with percentage change in accuracy we refer to Appendices~\ref{app:additional_tables} and~\ref{app:additional_figures}.

\begin{figure*}[t!]
    \centering
    \includegraphics[width=1\linewidth]{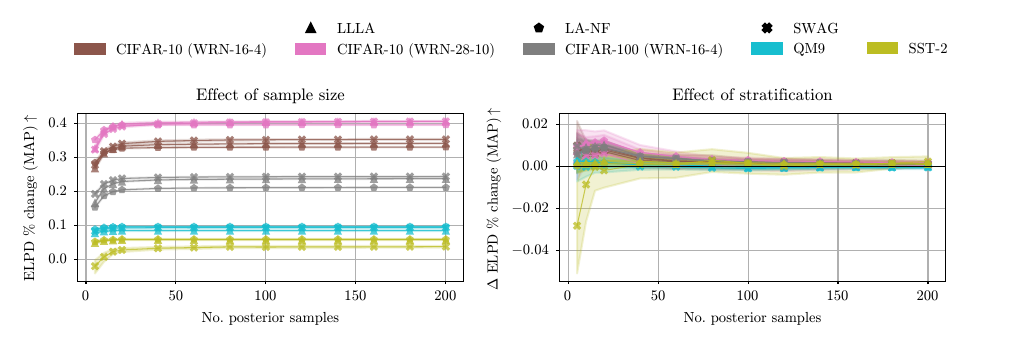}
    \caption{Left: Effect of MC samples in predictive posterior across datasets and DE-BNN methods with $K=20$. y-axis is ELPD percentage change for a given method relative to the MAP estimate with $K=1$. Right: Effect of stratification of samples across ensemble members for DE-BNN methods across datasets with $K=20$. y-axis is the difference in ELPD percentage change between stratified or non-stratified samples for a given method.}
    \label{fig:sample_effect}
\end{figure*}

\begin{figure*}[t!]
    \centering
    \includegraphics[width=1\linewidth]{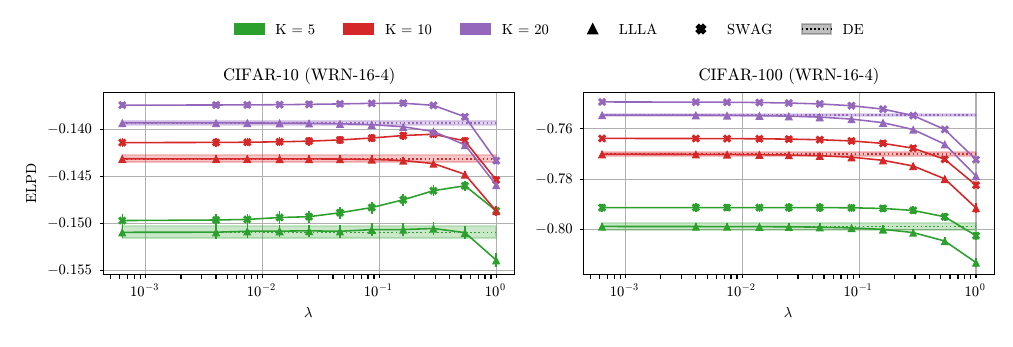}
    \caption{Test ELPD vs. covariance scaling factor $\lambda$ for WRN-16-4 on CIFAR-10 (left) and CIFAR-100 (right). Dotted lines indicate DE performance for a given K.}
    \label{fig:cov_scaling}
\end{figure*}

\subsection{DE-BNN Sensitivity Ablations}
\label{sec:sensitivity}
The previous experiment showed that DEs generally outperforms DE-BNNs for large ensembles sizes. In this section, we present a series of sensivity studies to shine more light on this observation.
First, it is natural to hypothesize that DE-BNNs with large $K$ requires a larger MC sample size for accurate results. 
Fig.~\ref{fig:sample_effect} shows the effect of increasing the number of MC samples used in the posterior predictive MC approximation and the effect of stratifying samples across the members in the DE-BNNs (for $K=20$). 

Firstly, Fig.~\ref{fig:sample_effect} (left) shows the ELPD percentage change over MAP estimated models as a function of MC samples. We observe across all approximate inference methods and datasets that the posterior predictive performance is saturated at $200$ MC samples. Hence, the MC sample size does not explain the fact that DEs outperforms DE-BNNs. 
Secondly, Fig.~\ref{fig:sample_effect} (right) shows the difference in ELPD percentage change for all DE-BNN methods when stratifying samples across the ensemble members. For all sample sizes the DE-BNNs perform marginally better with stratification, with the only outlier being SWAG on SST2.

Next, we investigate whether DE-BNNs can be improved by using non-uniform mixing weights. To study this, we estimate weights for each component using stacking \citep{yao2018stacking} based on the predictive performance on validation data in all of the ensembles with WRN-16-4 on CIFAR-10 and CIFAR-100. We found that that the resulting mixing weights were highly uniform: Computing the normalized entropy of the discrete distributions they represent, we obtain 0.99997 on CIFAR-10
%$0.99997\pm\num{1.92e-5}$ 
and 0.99999 on CIFAR-100
%$0.9999\pm\num{1.54e-4}$ 
with $1$ corresponding to a uniform distribution.

Since DEs can be interpreted as the limit of DE-BNNs, where the overall scale of the covariance matrices for each mixture component approaches zero, the last ablation study is designed to investigate the effect of shrinking the overall scale of covariance matrices uniformly for each component of the DE-BNNs. For this experiment, we study the WRN-16-4s for CIFAR-10 and CIFAR-100, and manipulate the covariance matrices for each mixture component by multiplying with a scalar $\lambda \in \left[10^{-3}, 1\right]$. Scaling with $\lambda = 10^{-3}$ effectively reduces SWAG to SWA and LLLA to MAP. Fig.~\ref{fig:cov_scaling} shows the results for SWAG and LLLA. It is seen that for all cases, except SWAG for $K=5$ on CIFAR-10, the scale $\lambda = 1$ is actually suboptimal and shrinking the overall scale of the covariance matrices for the DE-BNNs always results in improved ELPD. Interestingly, for SWAG on CIFAR-10, there exists values for $\lambda \in \left(10^{-3}, 1\right)$ such that the resulting $\lambda$-DE-BNNs outperforms both DEs and DE-BNNs in terms of ELPD---this is especially true for $K\in\{5, 10\}$ but the effect diminishes with the ensemble size.% and $K=10$. 

\begin{figure*}[t!]
    \centering
    \includegraphics[width=1\linewidth]{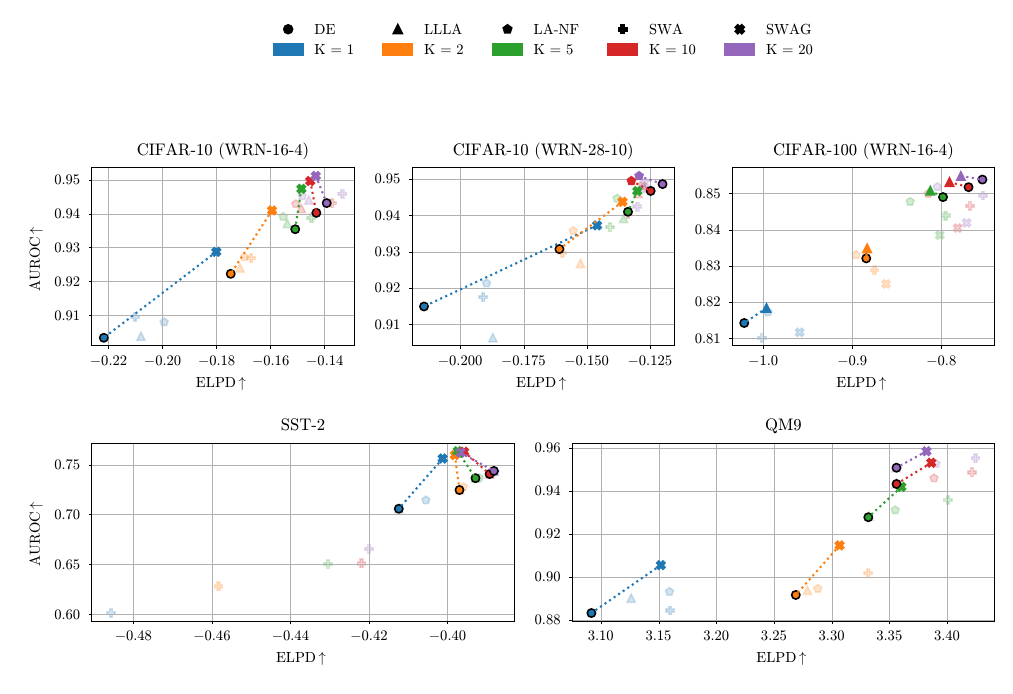}
    \caption{Out-of-distribution performance versus in-distribution test performance for DEs and DE-BNNs. DEs are highlighted with black marker edges. Lines are drawn from each DE to the DE-BNN that performs the best on the AUROC metric for a given $K$ and dataset. Some points are faded for clarity.}
    \label{fig:id_vs_ood}
\end{figure*}

\subsection{Out-of-Distribution Detection}
In the last experiment, we compare DEs and DE-BNNs in terms of out-of-distribution (OOD) detection.
 Fig.~\ref{fig:id_vs_ood} visualizes the AUROC \citep{hanley1982auroc} as a function of the test set ELPD (ID) for each model and dataset. It is seen that for small $K$, the DE-BNNs outperform the DEs, not just on in-distribution ELPD, but also on the AUROC. However, as the ensembles grow larger, i.e., for $K\in\{10, 20\}$, the DE-BNNs still outperforms the DEs w.r.t.\ the AUROC metric, but at the cost of reduced ID performance. Although it is observed that DE-BNNs constructed from SWAG models perform better than the MAP estimated DEs on QM9, we note that SWA DEs, another point estimate-based DE performs even better than the SWAG DE-BNNs. Lastly, the figure also shows that the best approximate inference method for OOD detection is dataset and model-dependent.

\section{DISCUSSION}

\subsection{DEs: The Practitioner's Choice}

We firstly discuss whether DEs or DE-BNNs generally are recommendable to use in practice with the current state of approximate inference in BNNs. As can be seen in Fig.~\ref{fig:ELPD_metrics}, across all of the datasets and models, we observed that for small $K$, DE-BNNs generally outperformed DEs both on ID and OOD metrics. Yet, as $K$ increases to $10$ and above, DEs outperformed DE-BNNs on ID metrics with DE-BNNs remaining best on OOD metrics although by a very small margin. With this finding in mind, we present two recommendations. 

Firstly, if a practitioner is limited on memory for their models at prediction time, it may be beneficial to consider sub-network BNNs such as LLLA due to the limited expense of memory in such sub-network approximations and the benefit afforded by them in regimes where it not feasible to use many ensemble members. Secondly, in the regime where the practitioner is not limited by memory and can increase their ensemble member size, then DEs appear to be the better choice not only due to predictive performance, but also due to their simplicity and efficiency in terms of implementation and fitting. 

Although DE-BNNs using SWAG often performed well, the memory cost and computational cost at prediction time of these models are often much larger than simply increasing the ensemble size due to how the low-rank covariance matrix is constructed and used. The memory footprint of a DE-BNN using SWAG is $\Theta(KPR)$, where $K$ is the number of members, $P$ is the number of parameters in the model and $R$ is the rank of the low-rank approximation in SWAG. In contrast, the footprint for DEs is simply $\Theta(KP)$. The computational cost at prediction time is generally also much lower for DEs over SWAG-based DE-BNNs. Moreover, some of our experiments (see e.g. Fig.~\ref{fig:ELPD_metrics} and Fig.~\ref{fig:cov_scaling}) also suggest that part of the gain observed for DE-BNNs with SWAG for larger $K$s can be explained by the mean of the approximate posterior distribution being moved towards $\theta_{\text{SWA}}$ rather than a gain due to local marginalization. See Appendix \ref{app:additional_tables} for a full overview of memory footprint, and training and prediction time cost for each model type.

\subsection{On Tuning DE-BNNs}
Although DE-BNNs at times outperform DEs, e.g., on OOD-detection for larger $K$ or ID ELPD for lower $K$, they require careful tuning of additional hyperparameters, a contrast to the simplicity of DEs. In this work, we have carefully tuned the prior variances for LLLA and LA-NF as well as the constant learning rate used in SWAG, where especially the latter is costly. Considering the marginal benefits in OOD performance, it may be difficult to justify the computational cost and implementation burden of DE-BNNs.

The prior variance and learning rate tuning of the individual LLLA, LA-NF, and SWAG models based on the validation ELPD can be seen as a form of calibration. Even though this improves the uncertainty quantification of the individual models (see Appendix \ref{app:additional_tables}), this might not be beneficial to the ensemble of these models, as it has been noted that an ensemble of calibrated models can lead to an under-confident ensemble \citep{wu2021should}. An interesting future line of work would be to investigate how to optimally combine BNNs in an ensemble instead of the näive approach taken in this work and previous DE-BNN works. Our experiment in Fig.~\ref{fig:cov_scaling} also indicates that this could be a viable direction, since the crude approach of adjusting a common scaling factor $\lambda$ for each ensemble member's covariance matrix, yields a DE-BNN better than the naïvely combined DE-BNN on CIFAR-10 for SWAG and $K\in\{5,10\}$.

\section{SUMMARY \& FUTURE WORK}
We have investigated the effect of modelling local posterior structure in DEs (DE-BNNs) through approximate inference methods. We systematically evaluated popular approximate inference methods for BNNs and observed that they improve on MAP estimators for different data modalities. Furthermore, we have shown that DE-BNNs are superior to DEs for small ensembles, but they are mostly superfluous as the ensembles grow. For large ensembles, DEs generally have better ID performance and are much easier to train and have lower computational cost at prediction time. 
In line with previous works \citep{wilson2020bayesian, Eschenhagen2021MixturesOL}, we observed that DE-BNNs improve OOD-detection.  However, in this work we shine a new light on DE-BNNs and have shown that for large ensembles, the improvement in OOD-detection comes at the cost of decreased ID-performance.

Our results leads one to question how this counter intuitive behaviour of large DEs outperforming large DE-BNNs can persist across datasets and inference methods. Under the hypothesis that marginalization w.r.t.\ the true posterior distribution yields competitive predictive performance in the ID setting, there may be several explanations. Firstly, SWAG, LLLA, and LA-NF are fairly crude approximations: SWAG and LLLA both enforce Gaussianity, whereas LLLA and LA-NF only capture the posterior uncertainty for the last layer of the network, and ignore the rest.  
Secondly, the members of the ensembles are tuned individually, where one may hypothesise that joint tuning is likely to give better performance. Other possible explanations include poorly specified weight space priors or lack of reparametrization invariance \citep{roy2024reparameterizationinvarianceapproximatebayesian}. Under this new light on large DE-BNNs, investigating new multi-modal posterior approximations, both theoretically and empirically is an interesting research avenue. %for Bayesian deep learning community

\subsubsection*{Acknowledgments}
The authors acknowledge support from the Novo Nordisk Foundation under grant no NNF22OC0076658 (Bayesian neural networks for molecular discovery). 

\bibliography{aistats2025onlocal}

\section*{Checklist}

 \begin{enumerate}

 \item For all models and algorithms presented, check if you include:
 \begin{enumerate}
   \item A clear description of the mathematical setting, assumptions, algorithm, and/or model. Yes.
   \item An analysis of the properties and complexity (time, space, sample size) of any algorithm. No (although the used methods have well established time and space complexities in the original papers).
   \item (Optional) Anonymized source code, with specification of all dependencies, including external libraries. Yes.
 \end{enumerate}

 \item For any theoretical claim, check if you include:
 \begin{enumerate}
   \item Statements of the full set of assumptions of all theoretical results. Not Applicable.
   \item Complete proofs of all theoretical results. Not Applicable.
   \item Clear explanations of any assumptions. Yes.
 \end{enumerate}

 \item For all figures and tables that present empirical results, check if you include:
 \begin{enumerate}
   \item The code, data, and instructions needed to reproduce the main experimental results (either in the supplemental material or as a URL). Yes.
   \item All the training details (e.g., data splits, hyperparameters, how they were chosen). Yes.
     \item A clear definition of the specific measure or statistics and error bars (e.g., with respect to the random seed after running experiments multiple times). Yes.
     \item A description of the computing infrastructure used. (e.g., type of GPUs, internal cluster, or cloud provider). Yes.
 \end{enumerate}

 \item If you are using existing assets (e.g., code, data, models) or curating/releasing new assets, check if you include:
 \begin{enumerate}
   \item Citations of the creator If your work uses existing assets. Yes.
   \item The license information of the assets, if applicable. Not Applicable.
   \item New assets either in the supplemental material or as a URL, if applicable. Yes.
   \item Information about consent from data providers/curators. Not Applicable.
   \item Discussion of sensible content if applicable, e.g., personally identifiable information or offensive content. Not Applicable.
 \end{enumerate}

 \item If you used crowdsourcing or conducted research with human subjects, check if you include:
 \begin{enumerate}
   \item The full text of instructions given to participants and screenshots. Not Applicable.
   \item Descriptions of potential participant risks, with links to Institutional Review Board (IRB) approvals if applicable. Not Applicable.
   \item The estimated hourly wage paid to participants and the total amount spent on participant compensation. Not Applicable.
 \end{enumerate}

 \end{enumerate}

%Stuff to setup appendix.
 \onecolumn
 \renewcommand*{\thesection}{\Alph{section}}
\renewcommand*{\thesubsection}{\thesection{.\arabic{subsection}}}
\renewcommand*{\thesubsubsection}{\Alph{subsubsection}.}
\setcounter{section}{0}
\renewcommand\thefigure{\thesection.\arabic{figure}}
\setcounter{figure}{0}
\renewcommand\thetable{\thesection.\arabic{table}}
\setcounter{table}{0}

\aistatstitle{On Local Posterior Structure in Deep Ensembles: \\
Supplementary Materials}
\thispagestyle{empty}
\section{TRAINING DETAILS}
\label{app:training_details}
\subsection{BERT Details}
\label{app:bert_details}
We use Huggingface's implementation of BERT and downscale the models. We set the hidden size to 256, have 4 layers, 8 attention heads, an intermediate size of 2048 and dropout rate to 0.5 in both the hidden and attention layers. We use a single fully connected layer as the classification layer. The remaining hyperparameters and architectural details are the default of the Hugging Face BERTConfig.

\subsection{Data Splits}
Across all the datasets we use the same seed (0) for splitting the training dataset into validation and training data. For CIFAR-10 and CIFAR-100 we always use 2,000 validation points. For SST-2 we use the pre-established validation dataset. For QM9 we use 10,000 points for validation and 10,831 points for testing. For CIFAR-10, CIFAR-100 and SST-2, we always test on the full pre-established test sets. For the OOD experiments, we use the test splits of the CIFAR-10, CIFAR-100, and SVHN datasets, 1,821 randomly selected datapoints from the Yahoo Finance News dataset (same size as the SST-2 test set), and 10,831 randomly selected points from the test split of the Tencent Alchemy dataset (same size as the QM9 test set). 

\subsection{MAP Training}
\label{app:MAP_training}
For all of the MAP models trained on CIFAR-10 and CIFAR-100, we train using the SGD optimizer with a learning rate of 0.1 and cosine annealing, momentum of 0.9 and a weight decay of \num{5e-4}. For the WRN-16-4 model we train for 100 epochs, whilst we for the large WRN-28-10 model train for 200 epochs. We always employ early-stopping based on validation ELPD. We always use data augmentation when training MAP models - in particular we use random horizontal flipping and random cropping. We refer to the GitHub repository for details on these augmentation. On both CIFAR-10 and CIFAR-100 we use a batch size of 128 during training.\\

For the MAP models on SST-2, we use the SGD optimizer with a weight decay of \num{3e-3}, momentum of 0.9 and a learning rate of 0.005 with cosine annealing. We train for 100 epochs, but once again early stop based on validation ELPD. We split the full SST-2 sentences into phrases, a commonly used data augmentation for the SST-2 dataset.\\

For training the PaiNN MAP models on QM9, we generally follow the approach taken by \citet{schutt2021equivariant}. However, as mentioned in Section \ref{sec:experimental_setup}, we adapt the PaiNN model to be compatible with a heteroschedastic Gaussian likelihood and train using the negative log-likelihood in contrast to \citet{schutt2021equivariant} who train with the mean squared error loss. We train the models for 650 epochs but use early stopping when we observe convergence in terms of the validation ELPD. We use the AdamW optimzer with a weight decay factor of \num{0.01} and an initial learning rate of \num{5e-4} that is cosine annealed.

\subsection{LLLA Training}
\label{app:LLLA_training}
As a general rule we do not use data augmentation during fitting of the LLLA as recommended for BNNs in \cite{izmailov2021bayesian} - we also empirically experiment with this and find that the models perform worse when data augmentation has been used. We generally use full Hessian matrices except for the WRN-28-10 model due to memory constraints, where we instead use a diagonal covariance matrix. We fit all of the LLLA models using Monte Carlo sampling in the posterior predictive and cross validate the prior precision after fitting the Hessian. We cross-validate using a grid of size 21, with evenly spaced prior precision values in the range [-4, 4] in log space. These parameters are used across all of the datasets.

\subsection{LA-NF Training}
\label{app:LA-NF_training}
Like with the LLLA models we do not use data augmentation during fitting of the LA-NF models, and again verify emperically that this is better than including data augmentation. When fitting our flows we use radial transforms and train for 20 epochs as is done in \cite{kristiadi2022posterior}. We use a varying number of transforms, specifically $T=\{1, 5, 10, 30\}$, and as a difference to \cite{kristiadi2022posterior}, we cross-validate our prior precision here as well, sweeping over the values $[1, 5, 10, 20, 30, 40, 50, 70, 90, 100, 125, 150, 175, 200, 500]$, yet never observe the prior precision values to be selected on the boundaries of the grid. We use the Adam optimizer and an initial learning rate of 0.001 with cosine annealing as is done in \cite{kristiadi2022posterior}. Finally, we employ early-stopping for the LA-NF models on QM9 based on the validation ELPD, as these were prone to overfitting.

\subsection{SWA and SWAG Training}

We run constant SGD from a pretrained MAP solution for 100 epochs\footnote{except for the WRN-28-10 model where we use 25 epochs.} where we collect a parameter sample $\theta_m$ after each epoch. We store all these parameter samples and estimate the empirical mean and covariance of these samples using the SWAG equations given in Section \ref{sec:background}.

We tune the learning rate by running SWAG for 21 constant learning rates in the interval [\num{1e-1}, \num{1e-4}]\footnote{for the PaiNN model on QM9 we use 30 learning rates in the interval  [\num{1e-6}, \num{1e-10}].} for each of the 30 MAP estimates. For each MAP estimate, we select the learning rate that results in the SWAG approximation with the highest validation ELPD as well as the learning rate that results in the SWA solution, $\theta_{\mathrm{SWA}}$, with the highest validation ELPD.

\subsection{MCDO Training}
We generally use the MAP hyperparameters for Monte Carlo dropout training, but cross-validate the dropout rate. This dropout rate is used both during training and inference. For all models, we sweep the dropout rate using the grid $[0.01,0.05, 0.1,0.2,0.3,0.4,0.5]$. Furthermore, we include $0.6$ in the grid for the WRN-28-10 models, extend the grid with $[0.6, 0.7, 0.8, 0.9]$ for the BERT models, and extend the grid with [\num{5e-7}, \num{1e-6}, \num{5e-6}, \num{1e-5}, \num{5e-5}, \num{1e-4}, \num{5e-4}, \num{1e-3}, \num{5e-3}] for the PaiNN models. During training, we early-stop based on the MAP solution as it is too expensive to evaluate with the full predictive distribution during training.

\subsection{IVON Training}
For IVON we generally use the MAP hyperparameters and the recommendations given by \citet{variational2024shen}. For the WRN-16-4 and WRN-28-10 models, we follow the settings given in Appendix C.2 of \citet{variational2024shen}. For the WRN-28-10 models, we also train for 400 epochs but saw no significant improvements compared to 200 epochs of training. For the BERT and PaiNN models, we use the MAP hyperparameters and set the IVON hyperparameters to $\beta_2 =0.99995$, $h_0=0.1$, and $\lambda=N$ as recommended in Section 3 of \citet{variational2024shen}. As with Monte Carlo Dropout training, we early stop based on the IVON mean as it is again too expensive to evaluate with the full predictive distribution during training.

\subsection{Computational Resources}
We train all models on an internal GPU-cluster with a mix of NVIDIA GPUs and only require a single GPU for each model training.

\section{Why Filter Norm Response and Not Batch Normalization}
\label{app:FRN_BN}
We here discuss why we use Filter Response Normalization (FRN) \citep{singh2020filter} and not Batch Normalization (BN) \citep{ioffe2015batch} in the Wide Residual Networks (WRN). Including BN is problematic in a Bayesian context both because it introduces dependencies between individual data points in the likelihood \citep{izmailov2021bayesian} and because the BN statistics need to be recomputed for each posterior sample \citep{maddox2019simple}, significantly increasing cost at prediction time. We therefore instead use FRN in place of BN, but for completeness provide results for the same models with BN for a subset of the inference methods here in Table \ref{tab:metric_table_bn} (the MAP, LLLA and LA-NF models) to show that the conclusions remain the same in terms of performance.

\section{MCDO and IVON Results \& Discussion}
\label{app:IVON_MCDO_Discussion}

\begin{figure*}[h!]
    \centering
    \vspace{-0.4cm}
    
{%
        \includegraphics[width=1\linewidth]{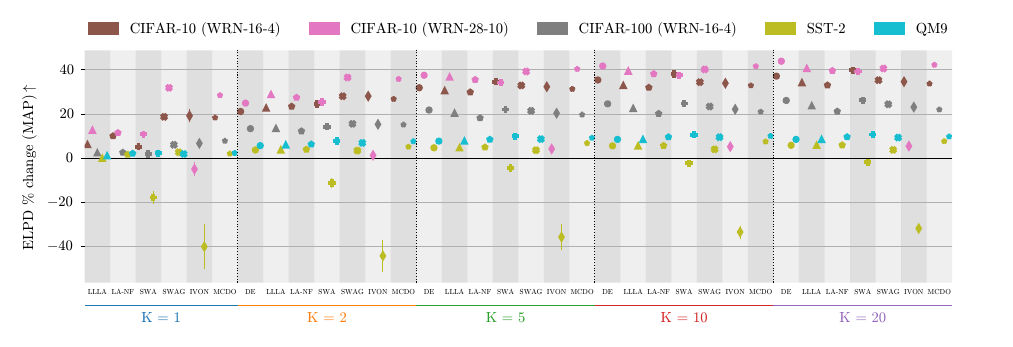}%
    }
    \vspace{-0.4cm}
    
{%
    \includegraphics[width=1\linewidth, trim={0 0 0 20}, clip]{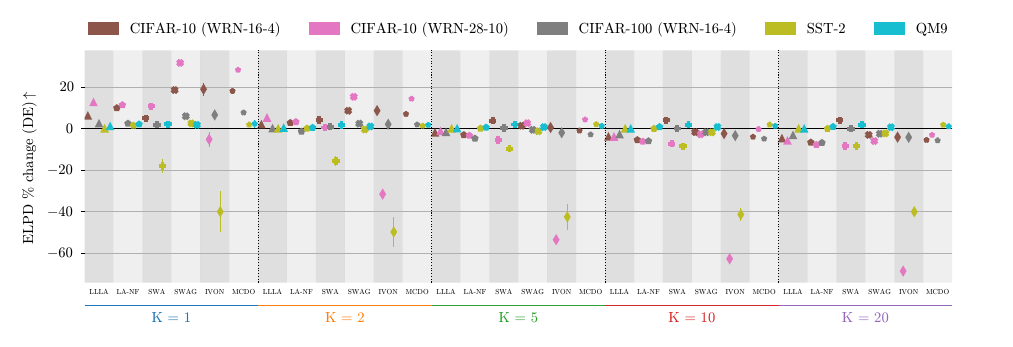}%
    }
    \caption{ELPD \% Change for All Inference Methods and Datasets Excluding IVON on QM9. (top) is versus MAP ($K=1$) and (bottom) is versus DE with same $K$. }
    \label{fig:elpd_metrics_wo_qm9_flow}
\end{figure*}

In this section, we discuss the IVON and MCDO results. For in-distribution ELPD performance we refer to Table \ref{tab:metric_table} and Figure \ref{fig:elpd_metrics_wo_qm9_flow}. For both IVON and MC-Dropout (MCDO), we generally observe the same pattern as with the remaining methods. For small $K$, DE-BNNs constructed from models trained with these methods outperform DEs, but as the ensembles grow large enough, the DEs outperform DE-BNNs on in-distribution metrics. 

We note that for the MCDO models on QM9 and SST-2, the DE-BNNs outperform DEs even for large $K$s, but this may be explained by the effect of better tuned regularization, due to increased dropout rates and more regularization in these models, rather than local marginalization. We also note that IVON models perform poorly even for small $K$ for a number of models, including the large WRN-28-10 on CIFAR-10 and the SST-2 and QM9 models. We hypothesize that this performance gap likely could be improved by more extensive hyperparameter tuning of the IVON optimizer. However, such hyperparameter tuning is highly expensive due to IVON training not being performed post-hoc, but rather when training from scratch. Please also note that the IVON models on QM9 have been left out of Figure \ref{fig:elpd_metrics_wo_qm9_flow} because these models have not converged to a meaningful solution as evident in Table \ref{tab:metric_table}.

\begin{figure*}[h!]
    \centering
    \includegraphics[width=1\linewidth]{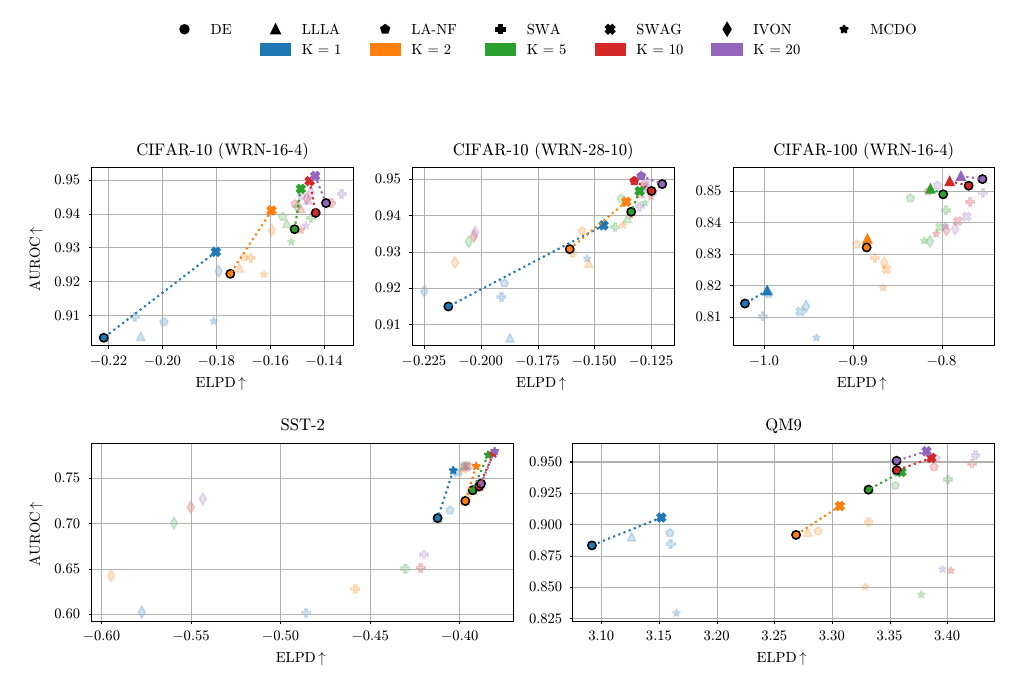}
    \caption{Out-of-Distribution Performance versus In-Distribution Test Performance for all DEs and DE-BNNs.}
    \label{fig:id_vs_ood_app}
\end{figure*}

With regards to OOD performance we refer to Figure \ref{fig:id_vs_ood_app}. Here we observe that consistent with the other methods, the DE-BNNs constructed with well-tuned IVON and MCDO models consistently outperform the DEs as previously noted.

\begin{table}[H]
\centering
\caption{WRN Results with Batch Normalization} 
%\resizebox{\textwidth}{!}{
\begin{tabular}{llrrrrrr}
\toprule
 &  & \multicolumn{3}{c}{CIFAR-10 (WRN-16-4)} & \multicolumn{3}{c}{CIFAR-100 (WRN-16-4)} \\
 \cmidrule(lr){3-5} \cmidrule(lr){6-8}
K & Inference  & Acc.$\uparrow$ & ELPD$\uparrow$ & ECE$\downarrow$ & Acc.$\uparrow$ & ELPD$\uparrow$ & ECE$\downarrow$ \\
%K & Inference &  &  &  &  &  &  \\
\midrule
\multirow[c]{6}{*}{1} & DE & \textbf{0.947} & -0.185 & 0.023 & \textbf{0.760} & -0.949 & 0.082 \\
 & LLLA & \textbf{0.947} & -0.174 & 0.014 & \textbf{0.760} & -0.885 & 0.029 \\
 & LLLA-NF-1 & \textbf{0.947} & -0.171 & 0.010 & \textbf{0.760} & \textbf{-0.884} & 0.022 \\
 & LLLA-NF-5 & \textbf{0.947} & \textbf{-0.169} & \textbf{0.008} & \textbf{0.760} & -0.886 & \textbf{0.019} \\
 & LLLA-NF-10 & \textbf{0.947} & \textbf{-0.169} & \textbf{0.008} & \textbf{0.760} & -0.886 & \textbf{0.019} \\
 & LLLA-NF-30 & \textbf{0.947} & \textbf{-0.169} & 0.009 & \textbf{0.760} & -0.893 & 0.027 \\
\cline{1-8}
\multirow[c]{6}{*}{5} & DE & \textbf{0.957} & \textbf{-0.134} & \textbf{0.005} & 0.800 & \textbf{-0.719} & \textbf{0.019} \\
 & LLLA & \textbf{0.957} & \textbf{-0.134} & 0.008 & 0.801 & -0.736 & 0.070 \\
 & LLLA-NF-1 & \textbf{0.957} & -0.135 & 0.012 & \textbf{0.802} & -0.753 & 0.087 \\
 & LLLA-NF-5 & \textbf{0.957} & -0.137 & 0.015 & \textbf{0.802} & -0.750 & 0.082 \\
 & LLLA-NF-10 & \textbf{0.957} & -0.137 & 0.015 & \textbf{0.802} & -0.750 & 0.082 \\
 & LLLA-NF-30 & \textbf{0.957} & -0.137 & 0.014 & \textbf{0.802} & -0.765 & 0.093 \\
\cline{1-8}
\multirow[c]{6}{*}{10} & DE & \textbf{0.959} & \textbf{-0.128} & \textbf{0.005} & 0.806 & \textbf{-0.688} & \textbf{0.025} \\
 & LLLA & \textbf{0.959} & -0.131 & 0.011 & 0.807 & -0.717 & 0.078 \\
 & LLLA-NF-1 & \textbf{0.959} & -0.132 & 0.015 & \textbf{0.808} & -0.737 & 0.096 \\
 & LLLA-NF-5 & \textbf{0.959} & -0.134 & 0.017 & \textbf{0.808} & -0.733 & 0.091 \\
 & LLLA-NF-10 & \textbf{0.959} & -0.133 & 0.017 & \textbf{0.808} & -0.733 & 0.091 \\
 & LLLA-NF-30 & \textbf{0.959} & -0.134 & 0.017 & 0.807 & -0.749 & 0.102 \\
\cline{1-8}
\multirow[c]{6}{*}{20} & DE & \textbf{0.959} & \textbf{-0.125} & \textbf{0.005} & 0.810 & \textbf{-0.672} & \textbf{0.028} \\
 & LLLA & \textbf{0.959} & -0.128 & 0.012 & 0.810 & -0.708 & 0.084 \\
 & LLLA-NF-1 & \textbf{0.959} & -0.130 & 0.015 & \textbf{0.811} & -0.729 & 0.102 \\
 & LLLA-NF-5 & \textbf{0.959} & -0.132 & 0.018 & \textbf{0.811} & -0.724 & 0.096 \\
 & LLLA-NF-10 & \textbf{0.959} & -0.131 & 0.018 & \textbf{0.811} & -0.723 & 0.096 \\
 & LLLA-NF-30 & \textbf{0.959} & -0.131 & 0.017 & \textbf{0.811} & -0.740 & 0.107 \\
\bottomrule
\end{tabular}
.
\label{tab:metric_table_bn}
%}
\end{table}

% \begin{table}[H]
% \centering
% \caption{CIFAR10 with BN and without augmentation} 
% \resizebox{\textwidth}{!}{
% \input{tables/results_cifar10_bn_aug_false}.
% \label{tab:cifar10_bn}
% }

% \end{table}

% \begin{table}[H]
% \centering
% \caption{CIFAR100 with BN and without augmentation} 
% \resizebox{\textwidth}{!}{
% \input{tables/results_cifar100_bn_aug_false}.
% \label{tab:cifar100_bn}
% }

%\end{table}
\newpage
\section{ADDITIONAL TABLES}
\label{app:additional_tables}
\begin{table*}[h!]
\label{tab:all_avg_rankings}
\centering
\caption{Average Rankings Across Datasets Including All Flow Lengths.} 
\resizebox{\textwidth}{!}{
\begin{tabular}{lrrrrrrrrrrrrrrr}
\toprule
& \multicolumn{3}{c}{K = 1} & \multicolumn{3}{c}{K = 2} & \multicolumn{3}{c}{K = 5} & \multicolumn{3}{c}{K = 10} & \multicolumn{3}{c}{K = 20} \\
\cmidrule(lr){2-4} \cmidrule(lr){5-7} \cmidrule(lr){8-10} \cmidrule(lr){11-13} \cmidrule(lr){14-16}
Inference & Acc. & ELPD & ECE & Acc. & ELPD & ECE & Acc. & ELPD & ECE & Acc. & ELPD & ECE & Acc. & ELPD & ECE \\

\midrule
DE & 7.0 & 8.9 & 8.2 & 6.0 & 7.4 & 5.4 & 4.9 & 4.2 & \textbf{3.1} & 4.8 & \textbf{3.5} & \textbf{3.3} & 4.6 & \textbf{3.7} & 3.6 \\
SWA & 3.4 & 7.1 & 8.6 & 3.3 & 5.3 & 7.0 & \textbf{4.5} & 4.3 & 4.4 & \textbf{4.7} & 4.2 & 3.4 & \textbf{4.5} & 4.2 & \textbf{3.3} \\
SWAG & \textbf{2.3} & \textbf{2.8} & 4.4 & \textbf{3.0} & 3.0 & 5.0 & \textbf{4.5} & 3.8 & 5.7 & 5.3 & 4.4 & 5.7 & 5.4 & 4.7 & 5.4 \\
LLLA & 7.3 & 6.5 & 6.2 & 6.1 & 5.7 & 4.4 & 5.3 & 5.5 & 5.2 & 5.3 & 5.4 & 5.5 & 5.0 & 5.5 & 6.0 \\
LLLA-NF-1 & 5.9 & 4.9 & 4.7 & 6.2 & 6.5 & 5.1 & 5.9 & 6.5 & 5.4 & 5.2 & 6.0 & 5.8 & 5.2 & 5.2 & 5.5 \\
LLLA-NF-5 & 5.5 & 5.0 & 4.2 & 5.6 & 5.6 & 5.1 & 5.2 & 6.3 & 5.8 & 4.8 & 6.2 & 5.9 & 5.2 & 6.1 & 5.5 \\
LLLA-NF-10 & 5.5 & 4.9 & \textbf{3.9} & 5.1 & 6.2 & 5.5 & 4.8 & 6.8 & 6.6 & \textbf{4.7} & 6.7 & 6.3 & 4.8 & 6.7 & 6.2 \\
LLLA-NF-30 & 5.8 & 5.5 & \textbf{3.9} & 5.3 & 5.7 & 5.2 & 4.8 & 6.6 & 6.2 & 4.8 & 7.0 & 6.4 & 4.9 & 7.1 & 6.4 \\
IVON & 7.4 & 6.6 & 6.7 & 8.4 & 7.0 & 8.2 & 8.7 & 7.6 & 8.4 & 8.8 & 7.8 & 8.4 & 8.6 & 7.9 & 8.5 \\
MCDO & 4.9 & \textbf{2.8} & 4.3 & 5.9 & \textbf{2.7} & \textbf{4.1} & 6.5 & \textbf{3.5} & 4.1 & 6.5 & 3.7 & 4.3 & 6.8 & 3.9 & 4.6 \\
\bottomrule
\end{tabular}

}
\end{table*}

\begin{table}[H]
\centering
\caption{Test Metric Means.} 
\resizebox{\textwidth}{!}{
\begin{tabular}{llrrrrrrrrrrrrrrr}
\toprule
 &  & \multicolumn{3}{c}{CIFAR-10 (WRN-16-4)} & \multicolumn{3}{c}{CIFAR-10 (WRN-28-10)} & \multicolumn{3}{c}{CIFAR-100 (WRN-16-4)} & \multicolumn{3}{c}{SST-2} & \multicolumn{3}{c}{QM9} \\
 \cmidrule(lr){3-5} \cmidrule(lr){6-8} \cmidrule(lr){9-11} \cmidrule(lr){12-14} \cmidrule(lr){15-17}
 K & Inference & Acc.$\uparrow$ & ELPD$\uparrow$ & ECE$\downarrow$ & Acc.$\uparrow$ & ELPD$\uparrow$ & ECE$\downarrow$ & Acc.$\uparrow$ & ELPD$\uparrow$ & ECE$\downarrow$ & Acc.$\uparrow$ & ELPD$\uparrow$ & ECE$\downarrow$ & MAE$\downarrow$ & ELPD$\uparrow$ & ECE$\downarrow$ \\
%K & Inference &  &  &  &  &  &  &  &  &  &  &  &  &  &  &  \\
\midrule
\multirow[c]{7}{*}{1} & DE & 0.938 & -0.222 & 0.032 & 0.947 & -0.215 & 0.033 & 0.736 & -1.022 & 0.066 & 0.818 & -0.412 & 0.033 & 8.859 & 3.091 & 0.051 \\
 & SWA & \textbf{0.942} & -0.210 & 0.031 & \textbf{0.953} & -0.191 & 0.030 & 0.739 & -1.002 & 0.058 & 0.811 & -0.486 & 0.089 & \textbf{8.115} & 3.160 & 0.041 \\
 & SWAG & 0.941 & -0.180 & 0.009 & 0.952 & \textbf{-0.146} & \textbf{0.006} & \textbf{0.744} & -0.960 & 0.020 & \textbf{0.825} & \textbf{-0.401} & 0.031 & 8.145 & 3.152 & 0.049 \\
 & LLLA & 0.938 & -0.208 & 0.018 & 0.947 & -0.187 & 0.016 & 0.735 & -0.997 & 0.025 & 0.818 & -0.412 & 0.033 & 8.914 & 3.126 & 0.047 \\
 & LA-NF & 0.938 & -0.199 & 0.013 & 0.947 & -0.190 & 0.015 & 0.735 & -0.995 & \textbf{0.013} & 0.823 & -0.406 & 0.027 & 8.553 & 3.159 & \textbf{0.034} \\
 & IVON & 0.940 & \textbf{-0.179} & \textbf{0.007} & 0.927 & -0.225 & 0.032 & 0.738 & -0.953 & 0.017 & 0.655 & -0.578 & 0.057 & 2540.630 & -7.496 & 0.166 \\
 & MCDO & 0.940 & -0.181 & 0.010 & 0.952 & -0.153 & 0.014 & 0.733 & \textbf{-0.942} & 0.018 & 0.819 & -0.404 & \textbf{0.024} & 8.339 & \textbf{3.165} & 0.045 \\
\cline{1-17}
\multirow[c]{7}{*}{2} & DE & 0.947 & -0.175 & 0.014 & 0.954 & -0.161 & 0.015 & 0.761 & -0.885 & \textbf{0.019} & 0.827 & -0.397 & \textbf{0.024} & 7.306 & 3.268 & \textbf{0.081} \\
 & SWA & \textbf{0.949} & -0.167 & 0.014 & \textbf{0.957} & -0.160 & 0.019 & 0.763 & -0.876 & 0.020 & 0.818 & -0.458 & 0.074 & \textbf{6.842} & \textbf{3.331} & 0.082 \\
 & SWAG & 0.948 & -0.160 & 0.009 & 0.956 & \textbf{-0.136} & \textbf{0.005} & \textbf{0.767} & \textbf{-0.863} & 0.037 & 0.827 & -0.398 & 0.032 & 6.869 & 3.306 & 0.089 \\
 & LLLA & 0.947 & -0.171 & 0.009 & 0.954 & -0.153 & 0.007 & 0.761 & -0.884 & 0.032 & 0.827 & -0.397 & 0.025 & 7.359 & 3.279 & 0.082 \\
 & LA-NF & 0.947 & -0.170 & 0.009 & 0.954 & -0.156 & 0.009 & 0.761 & -0.896 & 0.054 & \textbf{0.829} & -0.396 & \textbf{0.024} & 7.358 & 3.288 & 0.082 \\
 & IVON & 0.947 & \textbf{-0.159} & 0.011 & 0.934 & -0.212 & 0.041 & 0.760 & -0.865 & 0.030 & 0.691 & -0.595 & 0.121 & 2536.273 & -7.276 & 0.165 \\
 & MCDO & 0.946 & -0.162 & \textbf{0.008} & 0.956 & -0.138 & 0.008 & 0.754 & -0.866 & 0.032 & 0.828 & \textbf{-0.391} & 0.026 & 6.953 & 3.329 & 0.082 \\
\cline{1-17}
\multirow[c]{7}{*}{5} & DE & 0.951 & -0.151 & 0.008 & 0.958 & -0.134 & 0.008 & 0.782 & -0.799 & \textbf{0.034} & \textbf{0.832} & -0.393 & 0.022 & 6.229 & 3.331 & 0.116 \\
 & SWA & \textbf{0.953} & \textbf{-0.145} & \textbf{0.007} & 0.958 & -0.141 & 0.014 & 0.781 & \textbf{-0.796} & 0.035 & 0.819 & -0.430 & 0.058 & \textbf{5.969} & \textbf{3.400} & \textbf{0.112} \\
 & SWAG & 0.952 & -0.149 & 0.015 & 0.958 & -0.130 & 0.008 & \textbf{0.784} & -0.802 & 0.063 & 0.826 & -0.397 & 0.031 & 6.012 & 3.360 & 0.118 \\
 & LLLA & 0.951 & -0.154 & 0.012 & 0.958 & -0.136 & 0.012 & 0.782 & -0.813 & 0.064 & \textbf{0.832} & -0.393 & 0.022 & 6.271 & 3.332 & 0.116 \\
 & LA-NF & 0.952 & -0.155 & 0.018 & \textbf{0.959} & -0.138 & 0.014 & 0.781 & -0.836 & 0.086 & 0.831 & -0.392 & \textbf{0.019} & 6.215 & 3.355 & 0.114 \\
 & IVON & 0.952 & -0.150 & 0.019 & 0.938 & -0.205 & 0.049 & 0.774 & -0.814 & 0.054 & 0.766 & -0.560 & 0.156 & 2539.758 & -7.223 & 0.165 \\
 & MCDO & 0.950 & -0.152 & 0.012 & 0.958 & \textbf{-0.128} & \textbf{0.005} & 0.768 & -0.820 & 0.056 & 0.831 & \textbf{-0.384} & 0.029 & 6.022 & 3.377 & 0.116 \\
\cline{1-17}
\multirow[c]{7}{*}{10} & DE & 0.953 & -0.143 & 0.009 & \textbf{0.960} & \textbf{-0.125} & 0.007 & 0.788 & -0.770 & \textbf{0.044} & 0.835 & -0.389 & 0.022 & 5.748 & 3.356 & 0.127 \\
 & SWA & \textbf{0.954} & \textbf{-0.137} & \textbf{0.006} & 0.959 & -0.134 & 0.012 & 0.788 & \textbf{-0.768} & \textbf{0.044} & 0.822 & -0.422 & 0.052 & 5.612 & \textbf{3.421} & \textbf{0.123} \\
 & SWAG & 0.953 & -0.145 & 0.018 & 0.958 & -0.128 & 0.008 & \textbf{0.791} & -0.782 & 0.073 & 0.828 & -0.396 & 0.033 & 5.645 & 3.386 & 0.127 \\
 & LLLA & 0.953 & -0.149 & 0.016 & \textbf{0.960} & -0.130 & 0.015 & 0.788 & -0.792 & 0.077 & 0.835 & -0.389 & 0.022 & 5.780 & 3.356 & 0.127 \\
 & LA-NF & 0.953 & -0.151 & 0.021 & \textbf{0.960} & -0.133 & 0.017 & 0.788 & -0.816 & 0.097 & 0.834 & -0.389 & \textbf{0.018} & 5.722 & 3.388 & 0.125 \\
 & IVON & 0.953 & -0.146 & 0.021 & 0.940 & -0.203 & 0.051 & 0.779 & -0.795 & 0.063 & 0.797 & -0.550 & 0.182 & 2537.032 & -7.140 & 0.165 \\
 & MCDO & 0.951 & -0.149 & 0.014 & 0.959 & \textbf{-0.125} & \textbf{0.004} & 0.773 & -0.807 & 0.065 & \textbf{0.836} & \textbf{-0.381} & 0.034 & \textbf{5.584} & 3.403 & 0.127 \\
\cline{1-17}
\multirow[c]{7}{*}{20} & DE & 0.954 & -0.139 & 0.009 & \textbf{0.961} & \textbf{-0.120} & 0.007 & 0.792 & -0.755 & \textbf{0.049} & \textbf{0.837} & -0.388 & 0.024 & 5.495 & 3.356 & 0.134 \\
 & SWA & \textbf{0.955} & \textbf{-0.134} & \textbf{0.005} & 0.960 & -0.130 & 0.011 & 0.792 & \textbf{-0.754} & 0.050 & 0.821 & -0.420 & 0.051 & \textbf{5.410} & \textbf{3.424} & \textbf{0.129} \\
 & SWAG & 0.954 & -0.143 & 0.018 & 0.958 & -0.127 & 0.009 & \textbf{0.794} & -0.772 & 0.079 & 0.826 & -0.397 & 0.032 & 5.438 & 3.382 & 0.134 \\
 & LLLA & 0.954 & -0.146 & 0.017 & \textbf{0.961} & -0.127 & 0.017 & 0.792 & -0.779 & 0.082 & \textbf{0.837} & -0.388 & 0.023 & 5.517 & 3.355 & 0.134 \\
 & LA-NF & 0.954 & -0.148 & 0.022 & \textbf{0.961} & -0.130 & 0.018 & 0.791 & -0.805 & 0.103 & 0.835 & -0.388 & \textbf{0.018} & 5.470 & 3.390 & 0.132 \\
 & IVON & 0.954 & -0.145 & 0.023 & 0.940 & -0.203 & 0.052 & 0.782 & -0.785 & 0.067 & 0.802 & -0.543 & 0.183 & 2540.415 & -7.130 & 0.165 \\
 & MCDO & 0.952 & -0.147 & 0.015 & 0.959 & -0.124 & \textbf{0.004} & 0.776 & -0.797 & 0.068 & \textbf{0.837} & \textbf{-0.381} & 0.036 & \textbf{5.410} & 3.395 & 0.134 \\
\bottomrule
\end{tabular}

\label{tab:metric_table}
}
\end{table}

\begin{table}[H]
\label{tab:metric_table_sem}
\centering
\caption{Standard Errors of Test Metric Means.}
\resizebox{\textwidth}{!}{
\begin{tabular}{llrrrrrrrrrrrrrrr}
\toprule
 &  & \multicolumn{3}{c}{CIFAR-10 (WRN-16-4)} & \multicolumn{3}{c}{CIFAR-10 (WRN-28-10)} & \multicolumn{3}{c}{CIFAR-100 (WRN-16-4)} & \multicolumn{3}{c}{SST-2} & \multicolumn{3}{c}{QM9} \\
 \cmidrule(lr){3-5} \cmidrule(lr){6-8} \cmidrule(lr){9-11} \cmidrule(lr){12-14} \cmidrule(lr){15-17}
 K & Inference & Acc. & ELPD & ECE & Acc. & ELPD & ECE & Acc. & ELPD & ECE & Acc. & ELPD & ECE & MAE & ELPD & ECE \\
%K & Inference &  &  &  &  &  &  &  &  &  &  &  &  &  &  &  \\
\midrule
\multirow[c]{7}{*}{1} & DE & 0.000 & 0.001 & 0.000 & 0.001 & 0.001 & 0.000 & 0.001 & 0.002 & 0.001 & 0.002 & 0.003 & 0.002 & 0.123 & 0.018 & 0.004 \\
 & SWA & 0.000 & 0.001 & 0.000 & 0.000 & 0.001 & 0.000 & 0.001 & 0.002 & 0.001 & 0.002 & 0.006 & 0.004 & 0.087 & 0.019 & 0.001 \\
 & SWAG & 0.000 & 0.001 & 0.000 & 0.000 & 0.001 & 0.000 & 0.001 & 0.002 & 0.000 & 0.001 & 0.002 & 0.002 & 0.088 & 0.018 & 0.002 \\
 & LLLA & 0.000 & 0.001 & 0.001 & 0.001 & 0.001 & 0.000 & 0.001 & 0.002 & 0.001 & 0.002 & 0.003 & 0.002 & 0.121 & 0.014 & 0.003 \\
 & LA-NF & 0.000 & 0.001 & 0.000 & 0.001 & 0.001 & 0.000 & 0.001 & 0.002 & 0.001 & 0.001 & 0.002 & 0.002 & 0.098 & 0.014 & 0.002 \\
 & IVON & 0.001 & 0.003 & 0.000 & 0.001 & 0.003 & 0.001 & 0.001 & 0.006 & 0.002 & 0.026 & 0.022 & 0.003 & 5.669 & 0.087 & 0.001 \\
 & MCDO & 0.000 & 0.001 & 0.001 & 0.000 & 0.001 & 0.000 & 0.001 & 0.002 & 0.002 & 0.001 & 0.002 & 0.002 & 0.104 & 0.019 & 0.003 \\
\cline{1-17}
\multirow[c]{7}{*}{2} & DE & 0.000 & 0.001 & 0.000 & 0.000 & 0.001 & 0.000 & 0.001 & 0.001 & 0.000 & 0.001 & 0.001 & 0.002 & 0.074 & 0.012 & 0.001 \\
 & SWA & 0.000 & 0.001 & 0.000 & 0.000 & 0.001 & 0.000 & 0.001 & 0.002 & 0.000 & 0.001 & 0.005 & 0.003 & 0.057 & 0.010 & 0.001 \\
 & SWAG & 0.000 & 0.000 & 0.000 & 0.000 & 0.000 & 0.000 & 0.000 & 0.001 & 0.000 & 0.001 & 0.001 & 0.002 & 0.057 & 0.010 & 0.001 \\
 & LLLA & 0.000 & 0.001 & 0.000 & 0.000 & 0.001 & 0.000 & 0.001 & 0.001 & 0.001 & 0.001 & 0.001 & 0.002 & 0.072 & 0.011 & 0.001 \\
 & LA-NF & 0.000 & 0.001 & 0.000 & 0.000 & 0.001 & 0.000 & 0.000 & 0.001 & 0.001 & 0.001 & 0.001 & 0.001 & 0.076 & 0.012 & 0.002 \\
 & IVON & 0.000 & 0.001 & 0.000 & 0.001 & 0.001 & 0.001 & 0.000 & 0.001 & 0.001 & 0.024 & 0.015 & 0.010 & 3.070 & 0.057 & 0.000 \\
 & MCDO & 0.000 & 0.000 & 0.000 & 0.000 & 0.000 & 0.000 & 0.001 & 0.002 & 0.001 & 0.001 & 0.001 & 0.002 & 0.043 & 0.009 & 0.001 \\
\cline{1-17}
\multirow[c]{7}{*}{5} & DE & 0.000 & 0.000 & 0.000 & 0.000 & 0.000 & 0.000 & 0.000 & 0.001 & 0.000 & 0.001 & 0.001 & 0.001 & 0.030 & 0.006 & 0.001 \\
 & SWA & 0.000 & 0.000 & 0.000 & 0.000 & 0.001 & 0.000 & 0.001 & 0.001 & 0.001 & 0.001 & 0.003 & 0.002 & 0.024 & 0.005 & 0.001 \\
 & SWAG & 0.000 & 0.000 & 0.000 & 0.000 & 0.000 & 0.000 & 0.000 & 0.001 & 0.000 & 0.001 & 0.001 & 0.001 & 0.025 & 0.006 & 0.001 \\
 & LLLA & 0.000 & 0.000 & 0.000 & 0.000 & 0.000 & 0.000 & 0.000 & 0.001 & 0.001 & 0.001 & 0.001 & 0.001 & 0.030 & 0.006 & 0.001 \\
 & LA-NF & 0.000 & 0.000 & 0.000 & 0.000 & 0.000 & 0.000 & 0.000 & 0.001 & 0.001 & 0.001 & 0.001 & 0.001 & 0.031 & 0.007 & 0.001 \\
 & IVON & 0.000 & 0.000 & 0.000 & 0.000 & 0.001 & 0.000 & 0.000 & 0.001 & 0.000 & 0.013 & 0.012 & 0.007 & 2.418 & 0.040 & 0.000 \\
 & MCDO & 0.000 & 0.000 & 0.000 & 0.000 & 0.000 & 0.000 & 0.000 & 0.001 & 0.001 & 0.001 & 0.001 & 0.001 & 0.040 & 0.006 & 0.001 \\
\cline{1-17}
\multirow[c]{7}{*}{10} & DE & 0.000 & 0.000 & 0.000 & 0.000 & 0.000 & 0.000 & 0.000 & 0.000 & 0.000 & 0.001 & 0.000 & 0.001 & 0.022 & 0.005 & 0.000 \\
 & SWA & 0.000 & 0.000 & 0.000 & 0.000 & 0.000 & 0.000 & 0.000 & 0.001 & 0.000 & 0.001 & 0.002 & 0.002 & 0.020 & 0.004 & 0.000 \\
 & SWAG & 0.000 & 0.000 & 0.000 & 0.000 & 0.000 & 0.000 & 0.000 & 0.000 & 0.000 & 0.001 & 0.001 & 0.001 & 0.020 & 0.004 & 0.000 \\
 & LLLA & 0.000 & 0.000 & 0.000 & 0.000 & 0.000 & 0.000 & 0.000 & 0.001 & 0.001 & 0.001 & 0.000 & 0.001 & 0.022 & 0.005 & 0.000 \\
 & LA-NF & 0.000 & 0.000 & 0.000 & 0.000 & 0.000 & 0.000 & 0.000 & 0.000 & 0.001 & 0.000 & 0.000 & 0.001 & 0.019 & 0.004 & 0.000 \\
 & IVON & 0.000 & 0.000 & 0.000 & 0.000 & 0.000 & 0.000 & 0.000 & 0.000 & 0.000 & 0.002 & 0.006 & 0.005 & 1.527 & 0.021 & 0.000 \\
 & MCDO & 0.000 & 0.000 & 0.000 & 0.000 & 0.000 & 0.000 & 0.000 & 0.001 & 0.001 & 0.001 & 0.000 & 0.001 & 0.030 & 0.004 & 0.000 \\
\cline{1-17}
\multirow[c]{7}{*}{20} & DE & 0.000 & 0.000 & 0.000 & 0.000 & 0.000 & 0.000 & 0.000 & 0.000 & 0.000 & 0.000 & 0.000 & 0.001 & 0.011 & 0.002 & 0.000 \\
 & SWA & 0.000 & 0.000 & 0.000 & 0.000 & 0.000 & 0.000 & 0.000 & 0.000 & 0.000 & 0.000 & 0.001 & 0.001 & 0.010 & 0.001 & 0.000 \\
 & SWAG & 0.000 & 0.000 & 0.000 & 0.000 & 0.000 & 0.000 & 0.000 & 0.000 & 0.000 & 0.001 & 0.001 & 0.001 & 0.009 & 0.002 & 0.000 \\
 & LLLA & 0.000 & 0.000 & 0.000 & 0.000 & 0.000 & 0.000 & 0.000 & 0.000 & 0.000 & 0.000 & 0.000 & 0.001 & 0.010 & 0.002 & 0.000 \\
 & LA-NF & 0.000 & 0.000 & 0.000 & 0.000 & 0.000 & 0.000 & 0.000 & 0.000 & 0.000 & 0.000 & 0.000 & 0.001 & 0.009 & 0.002 & 0.000 \\
 & IVON & 0.000 & 0.000 & 0.000 & 0.000 & 0.000 & 0.000 & 0.000 & 0.000 & 0.000 & 0.001 & 0.003 & 0.003 & 0.807 & 0.013 & 0.000 \\
 & MCDO & 0.000 & 0.000 & 0.000 & 0.000 & 0.000 & 0.000 & 0.000 & 0.001 & 0.000 & 0.000 & 0.000 & 0.001 & 0.015 & 0.002 & 0.000 \\
\bottomrule
\end{tabular}

}
\end{table}

\begin{table}[H]
\label{tab:metric_table_la}
\centering
\caption{Test Metric Means For All Flow Lengths.} 
\resizebox{\textwidth}{!}{
\begin{tabular}{llrrrrrrrrrrrrrrr}
\toprule
 &  & \multicolumn{3}{c}{CIFAR-10 (WRN-16-4)} & \multicolumn{3}{c}{CIFAR-10 (WRN-28-10)} & \multicolumn{3}{c}{CIFAR-100 (WRN-16-4)} & \multicolumn{3}{c}{SST-2} & \multicolumn{3}{c}{QM9} \\
 \cmidrule(lr){3-5} \cmidrule(lr){6-8} \cmidrule(lr){9-11} \cmidrule(lr){12-14} \cmidrule(lr){15-17}
 K & Inference & Acc.$\uparrow$ & ELPD$\uparrow$ & ECE$\downarrow$ & Acc.$\uparrow$ & ELPD$\uparrow$ & ECE$\downarrow$ & Acc.$\uparrow$ & ELPD$\uparrow$ & ECE$\downarrow$ & Acc.$\uparrow$ & ELPD$\uparrow$ & ECE$\downarrow$ & MAE$\downarrow$ & ELPD$\uparrow$ & ECE$\downarrow$ \\
%K & Inference &  &  &  &  &  &  &  &  &  &  &  &  &  &  &  \\
\midrule
\multirow[c]{5}{*}{1} & LLLA & \textbf{0.938} & -0.208 & 0.018 & 0.947 & \textbf{-0.187} & 0.016 & 0.735 & -0.997 & 0.025 & 0.818 & -0.412 & 0.033 & 8.914 & 3.126 & 0.047 \\
 & LLLA-NF-1 & \textbf{0.938} & -0.206 & 0.018 & 0.947 & \textbf{-0.187} & 0.016 & \textbf{0.736} & \textbf{-0.995} & 0.014 & 0.822 & \textbf{-0.405} & \textbf{0.025} & 8.590 & 3.155 & 0.035 \\
 & LLLA-NF-5 & \textbf{0.938} & -0.201 & 0.014 & \textbf{0.948} & -0.189 & \textbf{0.015} & 0.735 & \textbf{-0.995} & \textbf{0.013} & 0.822 & \textbf{-0.405} & 0.026 & 8.577 & 3.158 & 0.034 \\
 & LLLA-NF-10 & \textbf{0.938} & \textbf{-0.199} & 0.013 & 0.947 & -0.190 & \textbf{0.015} & 0.735 & \textbf{-0.995} & \textbf{0.013} & \textbf{0.823} & -0.406 & 0.027 & 8.553 & 3.159 & 0.034 \\
 & LLLA-NF-30 & \textbf{0.938} & \textbf{-0.199} & \textbf{0.012} & 0.947 & -0.190 & \textbf{0.015} & 0.735 & -1.000 & 0.020 & 0.822 & -0.406 & 0.027 & \textbf{8.535} & \textbf{3.161} & \textbf{0.032} \\
\cline{1-17}
\multirow[c]{5}{*}{2} & LLLA & \textbf{0.947} & -0.171 & \textbf{0.009} & \textbf{0.954} & \textbf{-0.153} & \textbf{0.007} & 0.761 & \textbf{-0.884} & \textbf{0.032} & 0.827 & -0.397 & 0.025 & 7.359 & 3.279 & 0.082 \\
 & LLLA-NF-1 & \textbf{0.947} & -0.171 & \textbf{0.009} & \textbf{0.954} & -0.154 & 0.011 & \textbf{0.762} & -0.892 & 0.049 & 0.828 & -0.397 & 0.023 & 9.618 & 3.149 & 0.087 \\
 & LLLA-NF-5 & \textbf{0.947} & \textbf{-0.170} & \textbf{0.009} & \textbf{0.954} & -0.155 & 0.010 & 0.761 & -0.896 & 0.053 & \textbf{0.829} & \textbf{-0.395} & \textbf{0.021} & 7.375 & 3.286 & 0.084 \\
 & LLLA-NF-10 & \textbf{0.947} & \textbf{-0.170} & \textbf{0.009} & \textbf{0.954} & -0.156 & 0.009 & 0.761 & -0.896 & 0.054 & \textbf{0.829} & -0.396 & 0.024 & 7.358 & 3.288 & 0.082 \\
 & LLLA-NF-30 & \textbf{0.947} & \textbf{-0.170} & \textbf{0.009} & \textbf{0.954} & -0.155 & 0.009 & 0.761 & -0.906 & 0.065 & \textbf{0.829} & -0.396 & 0.024 & \textbf{7.170} & \textbf{3.316} & \textbf{0.079} \\
\cline{1-17}
\multirow[c]{5}{*}{5} & LLLA & 0.951 & \textbf{-0.154} & \textbf{0.012} & 0.958 & \textbf{-0.136} & \textbf{0.012} & \textbf{0.782} & \textbf{-0.813} & \textbf{0.064} & \textbf{0.832} & -0.393 & 0.022 & 6.271 & 3.332 & 0.116 \\
 & LLLA-NF-1 & \textbf{0.952} & \textbf{-0.154} & 0.013 & 0.958 & -0.137 & 0.014 & \textbf{0.782} & -0.831 & 0.083 & 0.830 & -0.393 & 0.020 & 7.232 & 3.289 & 0.116 \\
 & LLLA-NF-5 & \textbf{0.952} & -0.155 & 0.016 & \textbf{0.959} & -0.137 & 0.013 & 0.781 & -0.835 & 0.086 & 0.831 & \textbf{-0.392} & \textbf{0.019} & 6.246 & 3.349 & 0.114 \\
 & LLLA-NF-10 & \textbf{0.952} & -0.155 & 0.018 & \textbf{0.959} & -0.138 & 0.014 & 0.781 & -0.836 & 0.086 & 0.831 & \textbf{-0.392} & \textbf{0.019} & 6.215 & 3.355 & 0.114 \\
 & LLLA-NF-30 & \textbf{0.952} & -0.155 & 0.018 & \textbf{0.959} & -0.138 & 0.014 & 0.780 & -0.849 & 0.097 & 0.831 & \textbf{-0.392} & \textbf{0.019} & \textbf{6.136} & \textbf{3.379} & \textbf{0.111} \\
\cline{1-17}
\multirow[c]{5}{*}{10} & LLLA & \textbf{0.953} & \textbf{-0.149} & \textbf{0.016} & \textbf{0.960} & \textbf{-0.130} & \textbf{0.015} & \textbf{0.788} & \textbf{-0.792} & \textbf{0.077} & \textbf{0.835} & \textbf{-0.389} & 0.022 & 5.780 & 3.356 & 0.127 \\
 & LLLA-NF-1 & \textbf{0.953} & \textbf{-0.149} & 0.017 & \textbf{0.960} & -0.131 & 0.016 & \textbf{0.788} & -0.812 & 0.096 & 0.834 & \textbf{-0.389} & \textbf{0.018} & 5.996 & 3.353 & 0.127 \\
 & LLLA-NF-5 & \textbf{0.953} & -0.150 & 0.020 & \textbf{0.960} & -0.131 & 0.016 & \textbf{0.788} & -0.815 & 0.097 & 0.834 & \textbf{-0.389} & \textbf{0.018} & 5.756 & 3.380 & 0.125 \\
 & LLLA-NF-10 & \textbf{0.953} & -0.151 & 0.021 & \textbf{0.960} & -0.133 & 0.017 & \textbf{0.788} & -0.816 & 0.097 & 0.834 & \textbf{-0.389} & \textbf{0.018} & 5.722 & 3.388 & 0.125 \\
 & LLLA-NF-30 & \textbf{0.953} & -0.151 & 0.021 & \textbf{0.960} & -0.133 & 0.017 & 0.787 & -0.829 & 0.107 & 0.834 & \textbf{-0.389} & \textbf{0.018} & \textbf{5.704} & \textbf{3.405} & \textbf{0.123} \\
\cline{1-17}
\multirow[c]{5}{*}{20} & LLLA & \textbf{0.954} & \textbf{-0.146} & \textbf{0.017} & \textbf{0.961} & \textbf{-0.127} & \textbf{0.017} & \textbf{0.792} & \textbf{-0.779} & \textbf{0.082} & \textbf{0.837} & \textbf{-0.388} & 0.023 & 5.517 & 3.355 & 0.134 \\
 & LLLA-NF-1 & \textbf{0.954} & \textbf{-0.146} & 0.018 & \textbf{0.961} & -0.128 & \textbf{0.017} & \textbf{0.792} & -0.802 & 0.102 & 0.836 & \textbf{-0.388} & 0.019 & 5.565 & 3.364 & 0.134 \\
 & LLLA-NF-5 & \textbf{0.954} & -0.148 & 0.021 & \textbf{0.961} & -0.128 & \textbf{0.017} & 0.791 & -0.805 & 0.103 & 0.835 & \textbf{-0.388} & \textbf{0.018} & 5.490 & 3.383 & 0.132 \\
 & LLLA-NF-10 & \textbf{0.954} & -0.148 & 0.022 & \textbf{0.961} & -0.130 & 0.018 & 0.791 & -0.805 & 0.103 & 0.835 & \textbf{-0.388} & \textbf{0.018} & 5.470 & 3.390 & 0.132 \\
 & LLLA-NF-30 & \textbf{0.954} & -0.149 & 0.023 & \textbf{0.961} & -0.130 & 0.018 & 0.791 & -0.820 & 0.113 & 0.835 & \textbf{-0.388} & \textbf{0.018} & \textbf{5.462} & \textbf{3.406} & \textbf{0.130} \\
\bottomrule
\end{tabular}

}
\end{table}

\begin{table}[H]
\label{tab:metric_table_la_sem}
\centering
\caption{Standard Errors of Test Metric Means For All Flow Lenghts.} 
\resizebox{\textwidth}{!}{
\begin{tabular}{llrrrrrrrrrrrrrrr}
\toprule
 &  & \multicolumn{3}{c}{CIFAR-10 (WRN-16-4)} & \multicolumn{3}{c}{CIFAR-10 (WRN-28-10)} & \multicolumn{3}{c}{CIFAR-100 (WRN-16-4)} & \multicolumn{3}{c}{SST-2} & \multicolumn{3}{c}{QM9} \\
 \cmidrule(lr){3-5} \cmidrule(lr){6-8} \cmidrule(lr){9-11} \cmidrule(lr){12-14} \cmidrule(lr){15-17}
 K & Inference & Acc. & ELPD & ECE & Acc. & ELPD & ECE & Acc. & ELPD & ECE & Acc. & ELPD & ECE & MAE & ELPD & ECE \\
%K & Inference &  &  &  &  &  &  &  &  &  &  &  &  &  &  &  \\
\midrule
\multirow[c]{5}{*}{1} & LLLA & 0.000 & 0.001 & 0.001 & 0.001 & 0.001 & 0.000 & 0.001 & 0.002 & 0.001 & 0.002 & 0.003 & 0.002 & 0.121 & 0.014 & 0.003 \\
 & LLLA-NF-1 & 0.000 & 0.001 & 0.000 & 0.001 & 0.001 & 0.000 & 0.001 & 0.002 & 0.001 & 0.001 & 0.002 & 0.002 & 0.100 & 0.014 & 0.002 \\
 & LLLA-NF-5 & 0.000 & 0.001 & 0.000 & 0.001 & 0.001 & 0.000 & 0.001 & 0.002 & 0.000 & 0.001 & 0.002 & 0.002 & 0.098 & 0.014 & 0.002 \\
 & LLLA-NF-10 & 0.000 & 0.001 & 0.000 & 0.001 & 0.001 & 0.000 & 0.001 & 0.002 & 0.001 & 0.001 & 0.002 & 0.002 & 0.098 & 0.014 & 0.002 \\
 & LLLA-NF-30 & 0.000 & 0.001 & 0.000 & 0.001 & 0.001 & 0.000 & 0.001 & 0.002 & 0.001 & 0.001 & 0.002 & 0.002 & 0.096 & 0.015 & 0.001 \\
\cline{1-17}
\multirow[c]{5}{*}{2} & LLLA & 0.000 & 0.001 & 0.000 & 0.000 & 0.001 & 0.000 & 0.001 & 0.001 & 0.001 & 0.001 & 0.001 & 0.002 & 0.072 & 0.011 & 0.001 \\
 & LLLA-NF-1 & 0.000 & 0.001 & 0.000 & 0.000 & 0.001 & 0.000 & 0.000 & 0.001 & 0.002 & 0.001 & 0.001 & 0.002 & 0.627 & 0.030 & 0.005 \\
 & LLLA-NF-5 & 0.000 & 0.001 & 0.000 & 0.000 & 0.001 & 0.000 & 0.000 & 0.001 & 0.001 & 0.001 & 0.001 & 0.001 & 0.091 & 0.012 & 0.002 \\
 & LLLA-NF-10 & 0.000 & 0.001 & 0.000 & 0.000 & 0.001 & 0.000 & 0.000 & 0.001 & 0.001 & 0.001 & 0.001 & 0.001 & 0.076 & 0.012 & 0.002 \\
 & LLLA-NF-30 & 0.000 & 0.001 & 0.000 & 0.000 & 0.001 & 0.000 & 0.000 & 0.001 & 0.001 & 0.001 & 0.001 & 0.001 & 0.060 & 0.009 & 0.001 \\
\cline{1-17}
\multirow[c]{5}{*}{5} & LLLA & 0.000 & 0.000 & 0.000 & 0.000 & 0.000 & 0.000 & 0.000 & 0.001 & 0.001 & 0.001 & 0.001 & 0.001 & 0.030 & 0.006 & 0.001 \\
 & LLLA-NF-1 & 0.000 & 0.000 & 0.000 & 0.000 & 0.000 & 0.000 & 0.000 & 0.001 & 0.001 & 0.001 & 0.001 & 0.001 & 0.278 & 0.013 & 0.002 \\
 & LLLA-NF-5 & 0.000 & 0.000 & 0.000 & 0.000 & 0.000 & 0.000 & 0.000 & 0.001 & 0.000 & 0.001 & 0.001 & 0.001 & 0.030 & 0.007 & 0.001 \\
 & LLLA-NF-10 & 0.000 & 0.000 & 0.000 & 0.000 & 0.000 & 0.000 & 0.000 & 0.001 & 0.001 & 0.001 & 0.001 & 0.001 & 0.031 & 0.007 & 0.001 \\
 & LLLA-NF-30 & 0.000 & 0.000 & 0.000 & 0.000 & 0.000 & 0.000 & 0.000 & 0.001 & 0.001 & 0.001 & 0.001 & 0.001 & 0.023 & 0.006 & 0.001 \\
\cline{1-17}
\multirow[c]{5}{*}{10} & LLLA & 0.000 & 0.000 & 0.000 & 0.000 & 0.000 & 0.000 & 0.000 & 0.001 & 0.001 & 0.001 & 0.000 & 0.001 & 0.022 & 0.005 & 0.000 \\
 & LLLA-NF-1 & 0.000 & 0.000 & 0.000 & 0.000 & 0.000 & 0.000 & 0.000 & 0.001 & 0.001 & 0.001 & 0.000 & 0.001 & 0.068 & 0.005 & 0.000 \\
 & LLLA-NF-5 & 0.000 & 0.000 & 0.000 & 0.000 & 0.000 & 0.000 & 0.000 & 0.000 & 0.000 & 0.001 & 0.000 & 0.001 & 0.022 & 0.004 & 0.000 \\
 & LLLA-NF-10 & 0.000 & 0.000 & 0.000 & 0.000 & 0.000 & 0.000 & 0.000 & 0.000 & 0.001 & 0.000 & 0.000 & 0.001 & 0.019 & 0.004 & 0.000 \\
 & LLLA-NF-30 & 0.000 & 0.000 & 0.000 & 0.000 & 0.000 & 0.000 & 0.000 & 0.001 & 0.001 & 0.000 & 0.000 & 0.001 & 0.020 & 0.004 & 0.000 \\
\cline{1-17}
\multirow[c]{5}{*}{20} & LLLA & 0.000 & 0.000 & 0.000 & 0.000 & 0.000 & 0.000 & 0.000 & 0.000 & 0.000 & 0.000 & 0.000 & 0.001 & 0.010 & 0.002 & 0.000 \\
 & LLLA-NF-1 & 0.000 & 0.000 & 0.000 & 0.000 & 0.000 & 0.000 & 0.000 & 0.000 & 0.000 & 0.000 & 0.000 & 0.001 & 0.018 & 0.002 & 0.000 \\
 & LLLA-NF-5 & 0.000 & 0.000 & 0.000 & 0.000 & 0.000 & 0.000 & 0.000 & 0.000 & 0.000 & 0.000 & 0.000 & 0.001 & 0.009 & 0.002 & 0.000 \\
 & LLLA-NF-10 & 0.000 & 0.000 & 0.000 & 0.000 & 0.000 & 0.000 & 0.000 & 0.000 & 0.000 & 0.000 & 0.000 & 0.001 & 0.009 & 0.002 & 0.000 \\
 & LLLA-NF-30 & 0.000 & 0.000 & 0.000 & 0.000 & 0.000 & 0.000 & 0.000 & 0.000 & 0.000 & 0.000 & 0.000 & 0.000 & 0.009 & 0.002 & 0.000 \\
\bottomrule
\end{tabular}

}
\end{table}

\begin{table}[H]
\label{tab:probit}
\centering
\caption{Comparison of the Probit Approximation and Monte Carlo sampling in LLLA. The CV column denotes the approximation used during cross-validation (fitting) and the Test column denotes the approximation used during inference/test time.} 
%\resizebox{\textwidth}{!}{
\begin{tabular}{lllrrr}
\toprule
 &  & & \multicolumn{3}{c}{CIFAR-10 (WRN-16-4)} \\
 \cmidrule(lr){4-6}
 K & CV & Test & Acc.$\uparrow$ & ELPD$\uparrow$ & ECE$\downarrow$ \\
%K  & CV & Test &  &  &  \\
\midrule
\multirow[c]{3}{*}{1}  & MC & MC & \textbf{0.938} & -0.208 & 0.018 \\
  & MC & Probit & \textbf{0.938} & -0.358 & 0.146 \\
  & Probit & Probit & \textbf{0.938} & \textbf{-0.201} & \textbf{0.016} \\
\cline{1-6}
\multirow[c]{3}{*}{2}  & MC & MC & \textbf{0.947} & -0.171 & \textbf{0.009} \\
  & MC & Probit & 0.946 & -0.300 & 0.133 \\
  & Probit & Probit & 0.946 & \textbf{-0.170} & \textbf{0.009} \\
\cline{1-6}
\multirow[c]{3}{*}{5}  & MC & MC & 0.951 & \textbf{-0.154} & \textbf{0.012} \\
  & MC & Probit & 0.951 & -0.295 & 0.149 \\
  & Probit & Probit & \textbf{0.952} & \textbf{-0.154} & 0.017 \\
\cline{1-6}
\multirow[c]{3}{*}{10}  & MC & MC & \textbf{0.953} & \textbf{-0.149} & \textbf{0.016} \\
  & MC & Probit & \textbf{0.953} & -0.313 & 0.170 \\
  & Probit & Probit & \textbf{0.953} & \textbf{-0.149} & 0.020 \\
\cline{1-6}
\multirow[c]{3}{*}{20}  & MC & MC & \textbf{0.954} & \textbf{-0.146} & \textbf{0.017} \\
  & MC & Probit & \textbf{0.954} & -0.307 & 0.169 \\
  & Probit & Probit & \textbf{0.954} & -0.147 & 0.022 \\
\bottomrule
\end{tabular}

%}
\end{table}

\begin{table}[H]
\label{tab:scaling}
\centering
\caption{Memory, Prediction Time, and Training Cost in Big O Notation. In the table, $P_L$ is the number of parameters in the last layer, $P_{-L}$ is the number of parameters in the remaining layers, $P=P_{-L} + P_L$ is the total number of parameters,  $K$ is the number of ensemble members, $R$ is the rank of the low-rank SWAG approximation, $S$ is the number of samples used in the MC integration, $E$ is the number of epochs required during training to reach a mode, $F$ is the number of radial flows in LA-NF, and $E_V$ is the number of variational training epochs for LA-NF. Single mode approximations are special cases of the presented complexities where $K=1$.} 
%\resizebox{\textwidth}{!}{
\begin{tabular}{@{}llll@{}}
\toprule
Method     & Memory                   & Prediction Time           & Training Cost                     \\ \midrule
DE         & $KP$                     & $KP$                      & $KNPE$                            \\
Multi-SWAG & $KPR$                    & $SPR$                     & $KNP(E+R)$                        \\
MoLA       & $K(P_{-L}+P_L^2)$        & $KP_{-L}+SP_L^2$          & $KN(P_{-L}+P_L^3)+KNPE$           \\
Multi-IVON & $KP$                     & $SP$                      & $KNPE$                            \\
Multi-MCDO & $KP$                     & $SP$                      & $KNPE$                            \\
MoLA-NF    & $K(P_{-L}+P_L^2 + FP_L)$ & $KP_{-L}+S(P_L^2 + FP_L)$ & $KN(P_{-L}+P_L^3 + FP_LE_V)+KNPE$ \\ \bottomrule
\end{tabular}%
%}
\end{table}

\section{ADDITIONAL FIGURES}
\label{app:additional_figures}

\begin{figure*}[h!]
    \centering
    \vspace{-0.4cm}
    
{%
        \includegraphics[width=1\linewidth]{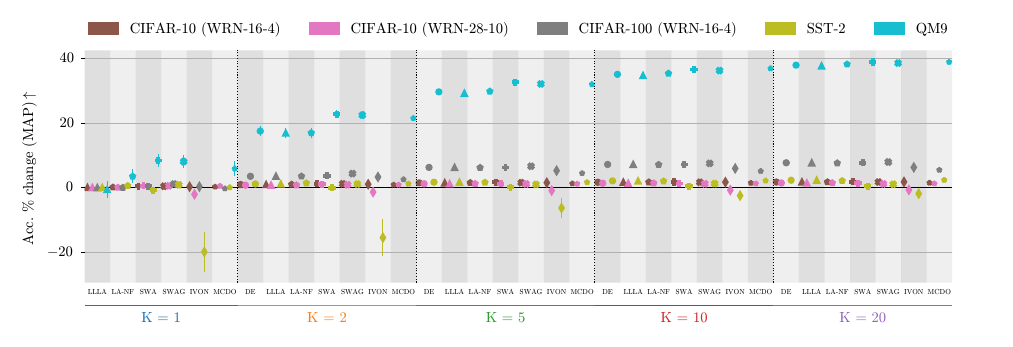}%
    }
    \vspace{-0.4cm}
    
{%
    \includegraphics[width=1\linewidth, trim={0 0 0 20}, clip]{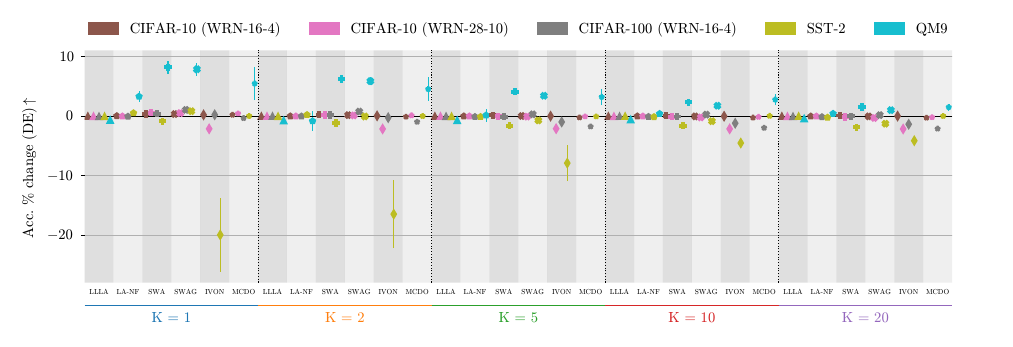}%
    }
    \caption{Accuracy \% Change for All Inference Methods and Datasets Excluding IVON on QM9. (top) is versus MAP ($K=1$) and (bottom) is versus DE with same $K$.}
    \label{fig:acc_metrics}
\end{figure*}

\begin{figure*}[h!]
    \centering
    \includegraphics[width=1\linewidth]{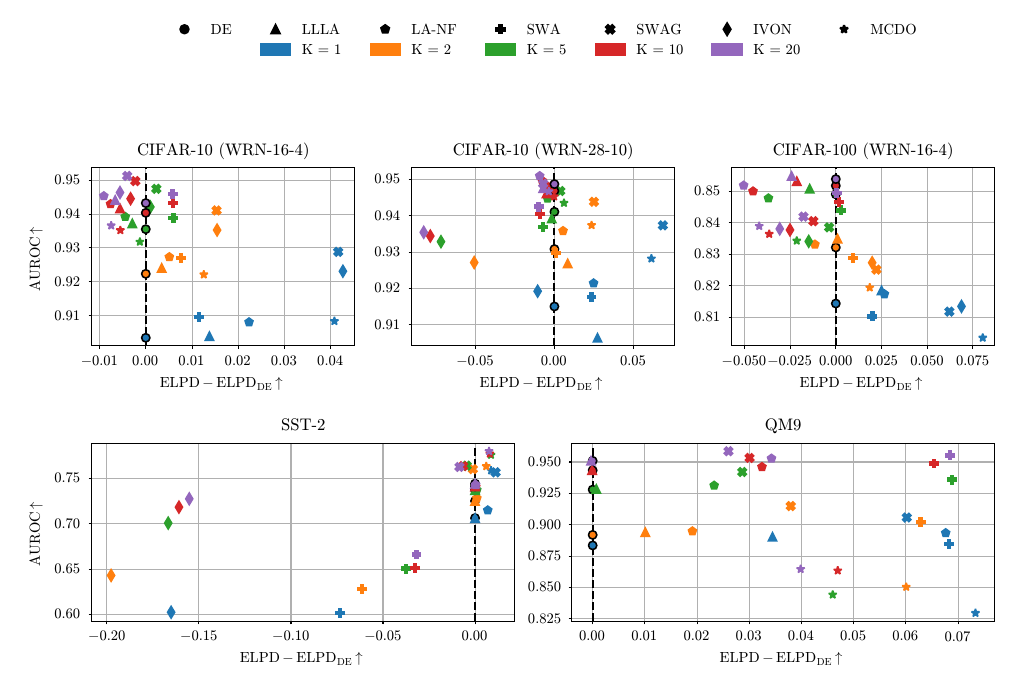}
    \caption{Out-of-Distribution Performance versus In-Distribution Test Performance for all DEs and DE-BNNs with x-axis being difference in ELPD versus DE.}
    \label{fig:id_vs_ood_2}
\end{figure*}

% \begin{figure*}[h!]
%     \centering
%     \includegraphics[width=1\linewidth]{figures/qm9_pred_var.pdf}
%     \caption{Average Predictive Variances for the QM9 Dataset.}
%     \label{fig:qm9_pred_var}
% \end{figure*}

\vfill
\end{document}